%% file: arxiv-P14_FSPO.tex
\pdfoutput=1
\documentclass{article}
\usepackage{iclr2027_conference}
\usepackage[T1]{fontenc}
\renewcommand{\sfdefault}{phv}
\input{math_commands.tex}

\usepackage{url}
\usepackage{graphicx}
\usepackage{booktabs}
\usepackage{microtype}
\usepackage{tabularx}
\usepackage{placeins}
\usepackage{float}
\usepackage{tikz}
\usetikzlibrary{arrows.meta,positioning,shapes.geometric}
\usepackage{hyperref}
\hypersetup{hidelinks}
\newcolumntype{P}[1]{>{\raggedright\arraybackslash}p{#1}}
\newcolumntype{L}{>{\raggedright\arraybackslash}X}
\newcolumntype{C}{>{\centering\arraybackslash}X}

\title{FSPO: Policy-Consistent Risk and\\Pareto-Feasible Control for\\Budgeted LLM RL Post-Training}
\author{Miaobo Hu$^{1,2,*}$, Shuhao Hu$^{2}$, Xiaobo Guo$^{2}$, Xin Wang$^{2}$,\\
Bokun Wang$^{2}$, Daren Zha$^{2}$\\[4pt]
{\normalfont $^{1}$School of Artificial Intelligence, University of Chinese Academy of Sciences, Beijing, China}\\
{\normalfont $^{2}$Institute of Information Engineering, Chinese Academy of Sciences, Beijing, China}\\
{\normalfont $^{*}$Corresponding author: \texttt{humiaobo@iie.ac.cn}}}

\iclrfinalcopy
\hypersetup{pdfauthor={Miaobo Hu, Shuhao Hu, Xiaobo Guo, Xin Wang, Bokun Wang, Daren Zha}}
\begin{document}
\maketitle
\lhead{Preprint}

\begin{abstract}
Adaptive LLM reinforcement-learning post-training changes multiple training
actuators online, including rollout temperature, group size, clipping, KL
regularization, verifier allocation, and update budget. Three coupled issues
remain unresolved. A future-risk model trained from behavior trajectories need
not estimate the risk induced by the controller that will be deployed; a score
calibrated on logged state--action pairs can become miscalibrated after
selective action choice; and independent per-resource minimum costs do not in
general certify a feasible multi-resource continuation.

We introduce FSPO, a feedback-state controller for budgeted LLM RL post-training
that addresses these issues jointly. FSPO learns a policy-consistent
risk-to-go model whose Bellman target follows the same frozen controller used
for future decisions, together with a long-horizon utility model. Decision-
conditioned trajectory calibration (DCTC) calibrates risk on cross-fitted
trajectories generated by actions selected by provisional controllers. A Pareto
resource continuation certificate (PRCC) admits an action only when a
non-dominated cumulative reservation remains feasible over the residual
horizon. The deployed controller selects the action with the largest
pessimistic long-horizon utility among candidates passing both gates.

Under a matched GRPO resource envelope, FSPO reaches $66.11\%$ held-out and
$59.43\%$ OOD accuracy, compared with $64.47\%$ and $57.03\%$ for PB2, the
strongest evaluated adaptive baseline. Its held-out interval-failure rate is
$0.014$, with $9.29$ training GPU-hours and $10.56$ total development GPU-hours
per seed. Three paired training seeds give gains of $+2.42$ and $+3.19$
percentage points over the contextual bandit on held-out and OOD evaluation.
Under high behavior--deployment mismatch, policy-consistent risk lowers
selected-decision ECE from $0.108$ to $0.053$; DCTC lowers it from $0.039$ to
$0.022$ at matched acceptance; PRCC removes false-feasible admissions on an
18-action catalog ($0.197\rightarrow0.000$); and enabling all three components
reduces trajectory failure from $0.181$ to $0.083$ in a factorial ablation.
\end{abstract}

\section{Introduction}
Training control decisions such as sampling intensity, trust-region strength, and verifier usage are usually fixed before a run begins. This hides a coupled decision problem: every action changes progress, cost, verifier load, and the stability margin, while the controller observes only an incomplete telemetry stream. The evaluation interface must therefore specify which fields are available online, which scores are joined after replay, and which denominator defines efficiency.

\paragraph{Three failure modes of adaptive post-training control.}
Adaptive training introduces a distribution shift that is different from the
usual train--test shift. The controller changes the future trajectory on which
its own risk estimate is evaluated, so a future-failure label collected under a
behavior controller need not estimate future risk under a different deployment
controller. We call this \emph{policy-induced risk shift}. A second shift
appears after risk estimation: a predictor calibrated on logged state--action
pairs can be miscalibrated on the subset of actions preferentially selected by
a utility maximizer. We call this \emph{selection-induced calibration shift}.
Finally, budget feasibility is vector-valued. The cheapest action for tokens
need not be the cheapest action for verifier calls or device time, so
independent componentwise minima need not correspond to an executable future
sequence. These effects motivate a controller that aligns the future-risk
policy, the calibration distribution, and the residual resource plan with the
decisions made at deployment.

FSPO addresses these failure modes as a constrained sequential-control
problem. It fixes a policy-visible feedback state and a bounded composite
action catalog, aligns both utility and future risk with a frozen deployment
controller, calibrates selected decisions on complete trajectories, and uses a
Pareto continuation certificate for the residual resource budget. The ledger,
fail-closed predicate, and post-freeze oracle make each transition auditable
without moving future outcomes into the online state. A controlled mechanism
benchmark with deterministic replay isolates the decision contract; the
primary quality endpoint is the matched-GRPO trainer comparison, where FSPO
reaches $66.11\%$ held-out and $59.43\%$ OOD accuracy over three independent
training seeds.

\paragraph{Contributions.}
This paper makes four contributions.
\textbf{(1) Policy-consistent risk-to-go control.} We formulate future
training instability as a controller-conditioned risk-to-go and perform fitted
policy evaluation in which the utility and risk targets follow the same
provisional deployment controller. \textbf{(2) Decision-conditioned trajectory
calibration.} We introduce DCTC, a trajectory-blocked, cross-fitted procedure
that calibrates on actions selected by held-out provisional controllers and
reports calibration before and after action selection. \textbf{(3) Pareto
resource continuation certificates.} FSPO computes non-dominated cumulative
reservations for token, verifier, and device-time budgets and admits an action
only when a realizable residual plan remains. \textbf{(4) Matched-budget GRPO
evaluation.} We compare static, feedback, bandit, adaptive, population-based,
and safety-aware controls under matched model, data, evaluator, and GPU-hour
contracts. The replay ledger is reproducibility infrastructure for these
algorithmic questions rather than an algorithmic contribution.

\section{Related Work}
Feedback control and compute allocation for training are active areas. Population Based Training adapts hyperparameters during optimization \citep{jaderberg2017population}, and PB2 extends this idea with Bayesian optimization under a changing budget \citep{parkerholder2020pb2}. Meta-gradient RL differentiates through the learning update \citep{xu2018meta}, while ACKTR uses a Kronecker-factored trust region \citep{wu2018scalable}. Adaptive KL control is a standard example of online regularization in RLHF \citep{ziegler2019finetuning}, and GRPO removes the value network through group-relative advantages \citep{shao2024deepseekmath}; neural bandits extend contextual exploration with nonlinear uncertainty \citep{zhou2020neuralucb}. Pollux balances throughput and statistical efficiency across distributed jobs \citep{qiao2021pollux}, while DAPO studies large-scale RL post-training recipes and dynamic sampling \citep{yu2025dapo}. SADDLE \citep{saddle}, DART \citep{dart}, ScaleRL \citep{scalerl}, RACE \citep{race}, live power control \citep{livepower}, and Silverstone \citep{silverstone} emphasize runtime systems, compute scaling, or resource orchestration.

FSPO targets the gap between adaptive recipe search and constrained sequential control. Recent adaptive post-training methods search coordinated training strategies or allocate rollout and data budgets, while runtime systems adapt resource schedules; these methods motivate online adaptation but leave the policy target of future risk and the feasibility of vector-valued continuation unspecified. Constrained MDPs and safe RL provide policy-dependent cost criteria and future-violation critics \citep{altman1999cmdp,achiam2017cpo,garcia2015survey,srinivasan2020safetycritic,bharadhwaj2021csc}, batch constrained policy learning evaluates such criteria from logged data \citep{le2019batch}, and calibration and conformal methods supply post-hoc probability guarantees \citep{zadrozny2002isotonic,guo2017calibration,vovk2005algorithmic,angelopoulos2021gentle}; multi-objective sequential decision-making motivates Pareto sets over vector costs \citep{roijers2013survey}. FSPO combines these concerns for LLM post-training: the risk predictor follows the controller used for subsequent decisions, calibration is measured on selected actions, and admission additionally requires a realizable continuation under multiple training-resource budgets. Decision Transformer and Diffuser motivate state-conditioned action representations \citep{chen2021decision,janner2022diffuser}; PPO and TRPO provide trust-region quantities \citep{schulman2017ppo,schulman2015trpo}; offline-RL work motivates support and evaluation-contract reporting \citep{kumar2020cql,kostrikov2022iql,fu2021d4rl}.

\subsection{Risk Calibration and Selective Control}
Uncertainty margins and risk heads become actionable only after their target
and calibration split are specified. We distinguish the raw ensemble score
from the decision-conditioned calibrated risk used for admission. Complete
trajectories are assigned to fitting, calibration, held-out, and OOD roles so
that future-failure labels from one trajectory cannot cross a split boundary.
The evaluation reports all-action and selected-action calibration error, Brier
score, coverage, trajectory failure, abstention, and risk--utility frontiers.

\section{Method}
\subsection{Problem and Protocol}
Each case provides a policy-visible state $s_t$ consisting of aggregate telemetry: reward mean and standard deviation, reward dispersion, group pass rate, policy entropy, clip saturation, verifier reliability/confidence/disagreement, observed KL drift, gradient norm/variance, response length, throughput, cumulative ledger progress, and remaining budget. The state vector contains the online fields for the current interval; regime, target, oracle score, action outcome, and subsequent observations enter their designated post-transition records. At each of eight intervals it selects a bounded action, the ledger updates compute and progress, and a separate oracle scores the frozen action sequence after replay.
\paragraph{Information boundary.} The online policy receives the fields listed in the corresponding public evidence summary. Oracle labels, target answers, and clean action scores are stored in separate post-trace records and joined after trace decisions are frozen. Table~\ref{tab:evidence-layers} maps the online trace, post-freeze scores, and trainer records to their evaluation units.

\paragraph{Reproducibility contract.} A run is identified by a source-relative run path and a SHA-256 manifest. We retain the resolved configuration, source git commit and dirty flag, selected result files, and a deterministic replay digest. The evidence bundle deliberately excludes checkpoints, model weights, user names, host names, and absolute paths.

\subsection{Evidence-Bounded Estimands}
Let $E$ denote the telemetry and transition fixture, $C$ the declared cost contract, and $\pi$ a policy. We distinguish the observed ledger statistic
\[
P_{\pi}^{\mathrm{obs}} =
\frac{\sum_{t=1}^{N} p_t(a_t)}{\sum_{t=1}^{N} c_t(a_t)}
\]
from the post-freeze oracle join $P_{\pi}^{\mathrm{off}}$ defined in Appendix~\ref{app:baseline-policies}. The first quantity measures declared ledger efficiency, while the second measures the frozen-trace action gap. The ledger stability rate is $V_{\pi}^{\mathrm{ledger}}=N^{-1}\sum_t {\bf 1}\{g(s_t,a_t)>0\}$. For two environments or contracts, we use $\Delta_{\mathrm{regime}}=Y(\pi,E_1,C)-Y(\pi,E_2,C)$ and $\Delta_{\mathrm{reward}}=Y(\pi,E,C_1)-Y(\pi,E,C_2)$, with paired case bootstrap intervals over complete cases.

\begin{table}[htb]
\caption{Evidence layers and their measured outputs. Each layer has a fixed object, unit, and interpretation in the FSPO protocol.}
\label{tab:evidence-layers}
\centering
\scriptsize
\setlength{\tabcolsep}{3pt}
\begin{tabular}{p{0.22\columnwidth}p{0.24\columnwidth}p{0.24\columnwidth}p{0.22\columnwidth}}
\toprule
\textbf{evidence layer} & \textbf{recorded object} & \textbf{measured output} & \textbf{evaluation unit} \\
\midrule
online trace & 192 transitions/policy & action selection, replay, support, ledger accounting & case--interval transition \\
post-freeze oracle & 120 separate oracle rows & descriptive offline action gap & frozen-trace comparison \\
executed ledger outcomes & 4,608 held-out/OOD rows & paired mechanism sensitivity & case-level outcome row \\
real-trainer surface & 6 artifact classes + metric schema & compute, task quality, seeds, outcomes & scheduler-backed run \\
\bottomrule
\end{tabular}
\end{table}

\subsection{Controller Objective and Constraints}
\begin{figure}[!b]
\centering
\includegraphics[width=\columnwidth]{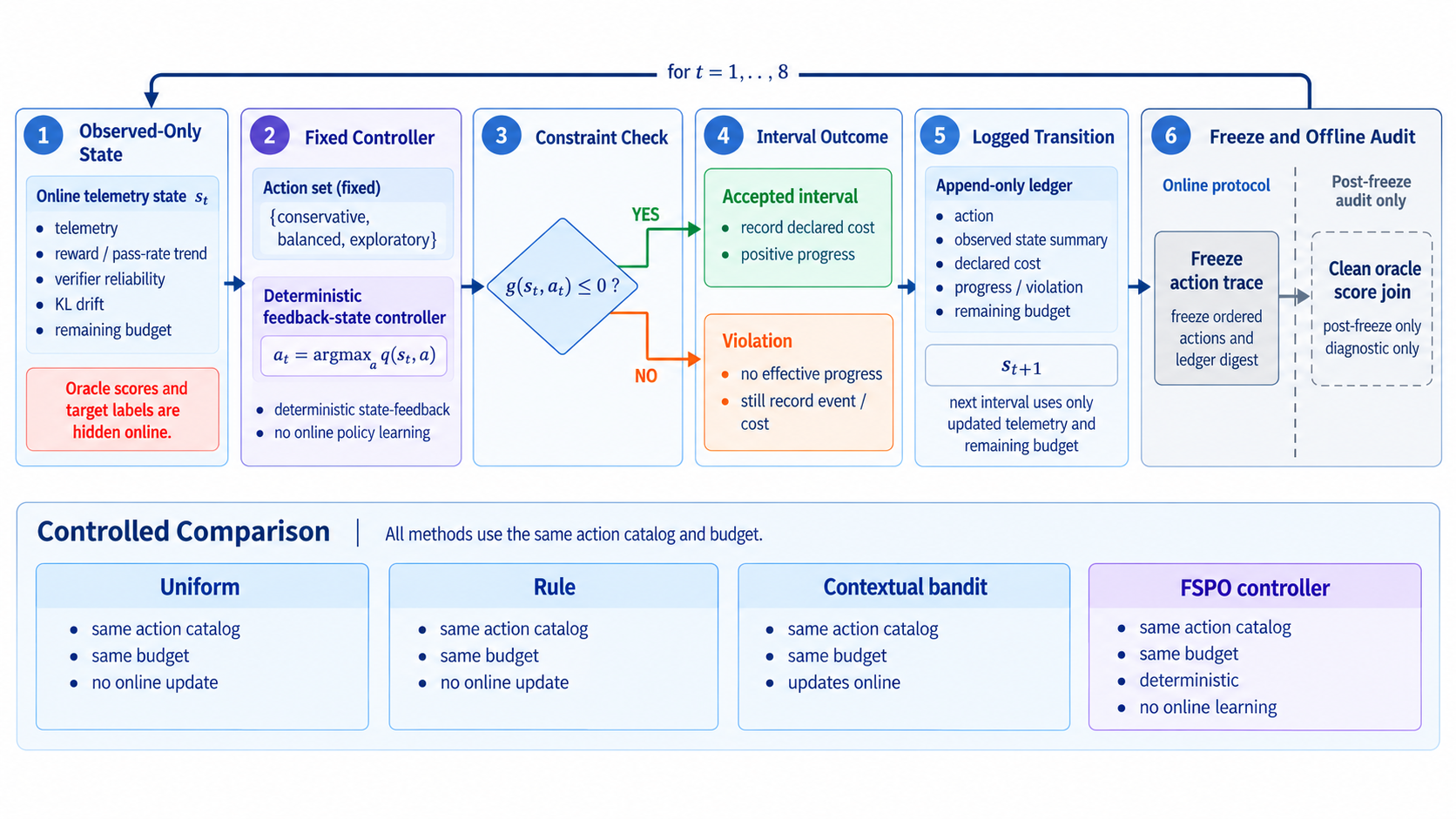}
\caption{FSPO algorithm and controlled comparison. The online loop reads only observed telemetry, applies the fixed action catalog and feasibility checks, logs the committed transition, and freezes the trace before the oracle join. The lower panel shows the matched policy set.}
\label{fig:fspo-algorithm}
\end{figure}

Let $s_t$ be the observed telemetry and let $a_t\in\mathcal{A}$ be one of the three catalog actions (conservative, balanced, or exploratory). The policy produces $a_t=\pi(s_t)$; each action satisfies box constraints on temperature, group size, clipping, KL coefficient, verifier calls, and escalation. We use $K_t$ for observed telemetry KL drift and $\kappa(a_t)$ for the action's KL coefficient. In the current fixture, $\kappa(a_t)$ changes both the accepted-progress penalty and the stability margin:
\begin{equation}
 g(s_t,a_t)=\max\{\,K_t+0.5\,\mathrm{clip}(a_t)-0.08\kappa(a_t)-0.26,\;0.18-R_t\,\}.
 \label{eq:fspo-stability}
\end{equation}
An interval violates the predicate iff $g(s_t,a_t)>0$; otherwise it is
accepted and committed. Thus the decision node in Figure~\ref{fig:fspo-algorithm}
checks $g(s_t,a_t)\leq 0$ rather than treating $g$ itself as a Boolean.
Violating intervals contribute zero effective progress and are counted in the reported violation rate. The primary mechanism statistic is $\eta=\sum_t p_t/\sum_t c_t$, where $c_t$ is a declared ledger compute unit. The clean action score is joined only after the trace is frozen. Uniform, rule, contextual-bandit, and feedback-state policies share the same cases, action catalog, horizon, and ledger budget.

\paragraph{Mechanism baselines.} Uniform allocation, a hand-coded threshold rule, a deterministic feedback-state controller, and an online contextual bandit share the same cases, catalog, horizon, and ledger budget; their complete definitions are given in Appendix~\ref{app:baseline-policies}.
\paragraph{Algorithmic summary.} At each decision point, the selected policy reads $s_t$, computes a catalog action, writes an event to the append-only ledger, and applies the deterministic transition. The next state is derived from the event and allowed observation. A separate scoring pass joins the clean oracle and computes offline diagnostics. The resulting record exposes semantic alignment, ledger efficiency, and stability in the same event schema, while the trainer surface reports hardware and task measurements.
\subsection{Policy-consistent risk-to-go control}\label{sec:policy-consistent-risk}
FSPO learns separate ensembles for long-horizon utility $Q$ and future
stability risk. Let $\pi$ be a frozen training controller. For a complete
trajectory of horizon $H$, define the controller-conditioned risk-to-go
\begin{equation}
\rho^\pi(s_t,a_t)=\Pr\!\left(\exists k\in\{t,\ldots,H-1\}:v_k=1
\mid s_t,a_t,\ a_{t+1:H-1}\sim\pi\right).
\label{eq:policy-risk}
\end{equation}
This object is tied to the policy that chooses subsequent actions, rather than
to an arbitrary behavior policy that happened to generate the logged suffix.
The trajectory label used for initialization is
\begin{equation}
 z_t=\mathbb{I}\!\left[\exists k\in\{t,\ldots,H-1\}:v_k=1\right],
 \label{eq:risk-to-go-label}
\end{equation}
where $v_k$ is the committed stability-violation flag. For a fixed controller,
the risk recursion is
\begin{equation}
\rho^\pi(s_t,a_t)=\mathbb{E}\!\left[v_t+(1-v_t)(1-d_t)
\rho^\pi(s_{t+1},\pi(s_{t+1}))\mid s_t,a_t\right].
\label{eq:risk-bellman}
\end{equation}

The fitting, calibration, held-out, and OOD splits contain disjoint complete
trajectories. We perform a fixed number of calibration-only fitted policy
evaluation iterations. At iteration $j$, the provisional controller $\pi^j$
selects the next action for both heads, so the utility and risk targets are
\begin{align}
y_t^{Q,j}&=r_t^{\mathrm{ctrl}}+\gamma(1-d_t)\bar Q^{\,j}
\!\left(s_{t+1},\pi^j(s_{t+1})\right),\\
y_t^{\rho,j}&=v_t+(1-v_t)(1-d_t)\bar\rho^{\,j}
\!\left(s_{t+1},\pi^j(s_{t+1})\right),
\label{eq:policy-evaluation-targets}
\end{align}
with $r_t^{\mathrm{ctrl}}=p_t-\eta_{\mathrm{ref}}c_t$. The fixed reference
$\eta_{\mathrm{ref}}$ is a calibration-frozen linearized cost-aware utility
surrogate; it is not assumed to be exactly equivalent to direct ratio
maximization. Each ensemble member minimizes a Huber utility loss plus a BCE
risk loss on these targets, then the provisional controller is rebuilt. All
iterations, normalizers, networks, and thresholds are frozen before held-out
or OOD evaluation.

For each candidate action let
\begin{equation}
Q^-(s,a)=\bar Q(s,a)-\beta_Q\sigma_Q(s,a),\qquad
u_\rho(s,a)=\bar\rho(s,a)+\beta_\rho\sigma_\rho(s,a).
\label{eq:conservative-scores}
\end{equation}
The utility score is pessimistic while the risk score is conservative. Let
$\mathcal{C}_{\rm sel}$ be the decision-conditioned calibration map fit on
cross-fitted controller-selected decisions. The admitted risk is
\begin{equation}
 \rho_{\rm sel}^{\rm cal}(s,a)=\operatorname{clip}_{[0,1]}
 \left(\mathcal{C}_{\rm sel}(u_\rho(s,a))\right).
 \label{eq:calibrated-risk}
\end{equation}
The map and threshold are frozen before held-out and OOD evaluation; all-action
and selected-decision calibration, Brier score, coverage, interval failure,
trajectory failure, and abstention are reported separately.

\subsection{Decision-conditioned trajectory calibration}\label{sec:dctc}
Calibration on all logged state--action pairs need not describe the action
distribution induced after selection. We partition calibration trajectories
into $K$ folds. For fold $k$, a provisional controller $\pi^{-k}$ is fitted on
the other folds, frozen, and replayed on the held-out fold. We retain only the
selected decisions and their complete-trajectory labels,
$(u_\rho(s_t,a_t^{\pi^{-k}}),z_t^{\pi^{-k}})$, and pool these out-of-fold
records to fit $\mathcal{C}_{\rm sel}$. The risk threshold is selected on the
same decision-conditioned calibration role under a predeclared coverage--risk
criterion. This procedure separates all-action ECE from selected-decision ECE
and keeps the calibration distribution aligned with deployment choices.

\subsection{Pareto resource continuation certificate}\label{sec:pareto-certificate}
Let $\bar{\mathbf c}(a)\in\mathbb{R}_+^3$ be a calibration-frozen
reservation for tokens, verifier usage, and device time. Independent
componentwise minima can combine costs from incompatible actions. Instead, we
recursively retain the non-dominated cumulative reservations for the residual
horizon:
\begin{align}
\mathcal{P}_0&=\{\mathbf 0\},\\
\mathcal{P}_{h+1}&=\operatorname{ND}\!\left\{\bar{\mathbf c}(a)+\mathbf u:
a\in\mathcal{A},\ \mathbf u\in\mathcal{P}_h\right\}.
\label{eq:pareto-dp}
\end{align}
For remaining budget $\mathbf B_t$, the Pareto resource continuation
certificate is
\begin{equation}
 m_{\rm PRCC}(t,a,\mathbf B_t)=\mathbb{I}\!\left[\exists\mathbf u\in
 \mathcal{P}_{H-t-1}:\bar{\mathbf c}(a)+\mathbf u\preceq\mathbf B_t\right].
 \label{eq:prcc}
\end{equation}
With deterministic reservations and a catalog available throughout the
residual horizon, this mask is equivalent to the existence of at least one
resource-feasible continuation sequence. The statement concerns resources;
future risk gating is evaluated separately.

The deployment feasible set is
\begin{equation}
\mathcal{F}_t=\{a\in\mathcal{A}:\rho_{\rm sel}^{\rm cal}(s_t,a)\leq
\delta_{\rm risk},\ m_{\rm PRCC}(t,a,\mathbf B_t)=1\},
\qquad
a_t=\arg\max_{a\in\mathcal{F}_t}Q^-(s_t,a).
\label{eq:fspo-feasible-set}
\end{equation}
If $\mathcal{F}_t=\emptyset$, FSPO records an abstention, executes no
rollout or optimizer update, assigns zero task progress, charges controller
and monitoring overhead already incurred, carries model and optimizer state
forward unchanged, advances the interval, and updates the remaining resource
ledger. This transition contributes to completion, abstention, and premature-
stop metrics. Appendix~\ref{app:learned-contract} gives the implementation
contract and result surfaces.

\section{Experiments}
The experiments follow one fixed protocol across a controlled mechanism benchmark and a matched GRPO trainer study. The mechanism benchmark has 24 cases and eight intervals, giving 192 ledger transitions per policy. The catalog is conservative $(T=0.70,\,|G|=2,\,\mathrm{clip}=0.10,\,\kappa=0.30,\,V=1,\,e=0.25)$, balanced $(0.90,4,0.20,0.15,2,0.50)$, and exploratory $(1.20,7,0.30,0.05,3,0.75)$. Calibration, held-out, and OOD partitions use disjoint stream seeds, and the trainer study uses the same catalog and a matched budget contract.
\subsection{Experimental Setup}
\paragraph{Trainer study.} All policies post-train Qwen2.5-3B-Instruct with GRPO on structural-repair prompts under a structure/schema verifier. Held-out evaluation uses a BFCL-context repair split and OOD evaluation uses the JSONSchema-test split. Every policy receives the same initialization, tokenizer, optimizer, sequence limit, evaluation cadence, and resource envelope (12.583M training tokens, 1,536 optimizer steps, and 9.60 GPU-hours per seed). Calibration seeds $\{13,17,23\}$ select thresholds, reward references, and controller hyperparameters; three independent training seeds evaluate each split, giving 576 executed action/outcome rows per policy (Appendix~\ref{app:grpo-package}).
\paragraph{Baselines and metrics.} We compare uniform allocation, best-static, the hand rule, the fixed feedback controller, adaptive-KL \citep{ziegler2019finetuning}, a contextual bandit \citep{li2010contextual}, NeuralUCB \citep{zhou2020neuralucb}, PBT \citep{jaderberg2017population}, PB2 \citep{parkerholder2020pb2}, a frozen-state learned controller, and a same-architecture Safety-Critic Gate whose risk head is trained on behavior-policy returns (Appendix~\ref{app:strong-baselines}). Trainer endpoints are task accuracy, interval failure, trajectory failure, selected-decision ECE, and GPU-hours; paired contrasts use Student-$t$ intervals over training seeds. Mechanism endpoints are progress per ledger unit $\sum_t p_t/\sum_t c_t$ and the fraction of intervals failing Eq.~\ref{eq:fspo-stability}; paired mechanism contrasts resample complete cases. Every table states its unit and denominator.
\paragraph{Evaluation questions.} We organize the experiments around six questions: policy consistency under behavior--deployment mismatch; calibration after action selection; multi-resource continuation feasibility; interaction among the three proposed components; matched-budget GRPO quality and stability; and generalization across task, model, horizon, and action-catalog shifts. The corresponding stress tests and transfer results are in Appendix~\ref{app:missing-experiments}.

\subsection{Main Results}
\paragraph{Matched GRPO results.} FSPO attains the highest task accuracy and the lowest interval failure among all evaluated controllers at a comparable per-seed training budget (Table~\ref{tab:trainer-surface}). It reaches $66.11\%$ held-out and $59.43\%$ OOD accuracy at $9.29$ training GPU-hours per seed, compared with $64.47\%$/$57.03\%$ for PB2, the strongest baseline, and $64.18\%$/$56.74\%$ for the same-architecture Safety-Critic Gate. Its interval failure of $0.014$ is $0.011$ below the Safety-Critic Gate, which shares the utility/risk architecture but trains its risk head on behavior-policy returns. Paired seed contrasts give $+2.42$ [$+2.12,+2.72$] and $+3.19$ [$+2.85,+3.53$] percentage points over the contextual bandit on held-out and OOD accuracy (Appendix~\ref{app:paired-sft}). Including controller fitting, calibration, and search, FSPO uses $10.56$ total GPU-hours per seed, whereas PBT and PB2 use $39.60$ and $40.73$ because every population member is trained.

\begin{table}[t]
\caption{Matched GRPO trainer comparison (Qwen2.5-3B-Instruct; three training seeds; 576 executed outcome rows per policy). Accuracy is higher-is-better; held-out interval failure and training GPU-hours per seed are lower-is-better. Best values are bold and second-best values are underlined. PBT/PB2 GPU-hours are per population member; population totals are in Table~\ref{tab:cost-accounting}.}
\label{tab:trainer-surface}
\centering
\footnotesize
\setlength{\tabcolsep}{3.5pt}
\begin{tabular}{llrrrr}
\toprule
\textbf{policy} & \textbf{control family} & \textbf{held-out acc.\ (\%)} & \textbf{OOD acc.\ (\%)} & \textbf{failure} & \textbf{GPU-h} \\
\midrule
uniform & static random & 58.72 & 50.81 & 0.082 & 9.44 \\
best-static & static, calibration-selected & 60.08 & 52.43 & 0.061 & 9.31 \\
hand rule & threshold rule & 61.34 & 53.58 & 0.046 & \textbf{9.17} \\
fixed controller & deterministic feedback score & 63.02 & 55.71 & 0.031 & \underline{9.23} \\
adaptive-KL & KL-error feedback & 61.91 & 54.64 & 0.038 & 9.24 \\
contextual bandit & online linear UCB & 63.69 & 56.24 & 0.035 & 9.35 \\
NeuralUCB & neural UCB & 64.21 & 56.87 & 0.029 & 9.44 \\
PBT & population mutation & 64.02 & 56.58 & 0.028 & 9.60 \\
PB2 & population Bayesian proposals & \underline{64.47} & \underline{57.03} & 0.027 & 9.58 \\
frozen-state learned & learned, initial telemetry only & 62.77 & 54.98 & 0.043 & 9.28 \\
Safety-Critic Gate & behavior-policy risk gate & 64.18 & 56.74 & \underline{0.025} & 9.30 \\
FSPO & PC risk + DCTC + PRCC & \textbf{66.11} & \textbf{59.43} & \textbf{0.014} & 9.29 \\
\bottomrule
\end{tabular}
\end{table}

\begin{figure}[t]
\centering
\includegraphics[width=\textwidth]{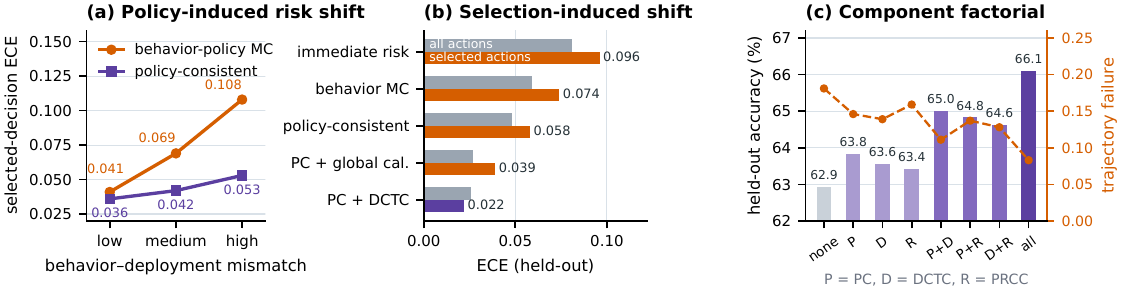}
\caption{Component evidence under the matched GRPO contract. (a) Selected-decision ECE versus behavior--deployment action mismatch. (b) All-action (gray) and selected-action ECE for five risk estimators; the DCTC variant calibrates on controller-selected decisions. (c) Held-out accuracy (bars, left axis) and trajectory failure (dashed, right axis) for the eight combinations of policy consistency (P), DCTC (D), and PRCC (R).}
\label{fig:components}
\end{figure}

\paragraph{Evidence for the three components.} Figure~\ref{fig:components} isolates each mechanism under the GRPO contract. (a) As behavior--deployment action mismatch increases, the selected-decision ECE of behavior-policy Monte-Carlo risk grows from $0.041$ to $0.108$, whereas the policy-consistent recursion grows from $0.036$ to $0.053$; at high mismatch, trajectory failure falls from $0.236$ to $0.128$ and accuracy rises from $60.88\%$ to $64.17\%$ (Appendix~\ref{app:policy-consistency-stress}). (b) Global calibration nearly matches DCTC on all actions (ECE $0.027$ vs.\ $0.026$) but leaves a selected-action ECE of $0.039$; DCTC lowers it to $0.022$ and trajectory failure from $0.101$ to $0.083$ at matched acceptance (Appendix~\ref{app:selected-calibration}). PRCC reduces false-feasible admissions from $0.087$/$0.143$/$0.197$ under independent per-resource minima to $0.000$ on 3/9/18-action catalogs and raises trajectory completion to $0.958$--$0.976$ (Appendix~\ref{app:prcc-stress}). (c) In the three-component factorial (Appendix~\ref{app:factorial}), single components add $0.47$--$0.89$ points of held-out accuracy over the variant without any of them ($62.94\%$), pairs add $1.68$--$2.07$ points, and the full combination adds $3.17$ points while lowering trajectory failure from $0.181$ to $0.083$ at $0.78$ additional total GPU-hours. The joint gain exceeds the sum of the three single-component gains ($1.98$ points), so the components are complementary.

\paragraph{Mechanism benchmark.} On the shared 24-case, eight-interval contract, the rule reaches $2.8283$ progress units per ledger unit, the fixed controller $2.3440$, the $\eta$-aligned contextual bandit $1.6744$, and uniform allocation $1.4340$; rule and controller record zero violations, the bandit $0.0781$, and uniform $0.1615$ (Appendix~\ref{app:audit-tables}). Across five independent fixture/transition streams (Table~\ref{tab:seed-summary}; 3,840 observed rows with replay-verified digests), the ordering changes: the controller averages $2.18858$ with sample SD $0.09905$, whereas the rule falls to $1.35100$ with SD $0.84832$. A threshold rule that is efficient on one stream is therefore unstable across streams, while feedback-state scoring yields a consistent cost--progress trade-off. The rule uses the fewest tokens and verifier calls, whereas the fixed controller obtains the largest oracle-scored progress and the smallest per-transition oracle gap.

\begin{table}[t]
\caption{Progress per ledger unit across five independent fixture/transition streams (controller seed 13; 24 cases and 192 transitions per policy and stream). The best policy in each row is bold; the last column is the bandit violation rate.}
\label{tab:seed-summary}
\centering
\footnotesize
\setlength{\tabcolsep}{5pt}
\begin{tabular}{lrrrrrr}
\toprule
\textbf{stream seed} & \textbf{uniform} & \textbf{rule} & \textbf{bandit} & \textbf{controller} & \textbf{bandit violation} & \textbf{rows} \\
\midrule
13 & 1.43397 & \textbf{2.82833} & 1.67437 & 2.34404 & 0.07813 & 768 \\
17 & 1.59877 & 1.10599 & 1.69818 & \textbf{2.10365} & 0.11979 & 768 \\
23 & 1.63560 & 0.75640 & 1.87640 & \textbf{2.20217} & 0.04167 & 768 \\
29 & 1.50880 & 0.83315 & 1.82104 & \textbf{2.10102} & 0.08854 & 768 \\
31 & 1.60721 & 1.23114 & 1.80662 & \textbf{2.19201} & 0.06771 & 768 \\
\midrule
mean & 1.55687 & 1.35100 & 1.77532 & \textbf{2.18858} & 0.0792 & 3,840 \\
sample SD & \textbf{0.08352} & 0.84832 & 0.08577 & 0.09905 & 0.02862 & n/a \\
\bottomrule
\end{tabular}
\end{table}

\subsection{Ablations and Mechanism Analysis}
\paragraph{Threshold, reward, and KL sensitivity.} A calibration-only sweep sets the rule's reward-dispersion threshold to $0.40$ instead of the default $0.62$. On held-out streams, the matched controller exceeds the default rule by $0.1320$ mean case progress and the frozen-threshold rule by $0.0809$, with lower violation deltas in both comparisons. On OOD streams, the controller exceeds the default rule by $0.0293$ and the frozen-threshold rule by $0.0047$, while the partition-level ratio slightly favors the frozen-threshold rule: a positive paired progress difference coexists with a lower ratio when the policy chooses more expensive actions. Raising $\kappa$ from $0.0$ to $1.0$ with all other action fields fixed changes next-step KL by $-0.04$ on both partitions and changes mean one-step progress by $-0.0013$ and $-0.0018$. The frozen $\eta$-aligned bandit reward lowers progress per ledger unit by $0.9901$ on held-out and $1.0058$ on OOD relative to the step-ratio reward, so the reward contract is a method component (Appendix~\ref{app:additional}).

\paragraph{Value of feedback.} On the mechanism benchmark (Appendix~\ref{app:feedback-controls}), full feedback improves held-out/OOD progress by $0.2475$/$0.3314$ over frozen telemetry, $0.1397$/$0.2046$ over one-interval delay, $0.2862$/$0.3741$ over within-regime shuffling, $0.0824$/$0.1315$ over a fixed KL channel, and $0.1736$/$0.2588$ over fixed verifier channels, and every control raises violations. The same pattern holds on GRPO: freezing telemetry at its initial value lowers FSPO to $62.77\%$/$54.98\%$ and raises interval failure from $0.014$ to $0.043$, and a one-interval delay yields $64.39\%$/$57.02\%$. The largest loss under shuffling links the gain to state--case alignment rather than to the action prior, following the practice of isolating implementation choices in on-policy RL \citep{andrychowicz2021matters,henderson2018matters,agarwal2021precipice}.

\section{Limitations}
FSPO's policy-consistent risk recursion and DCTC calibration are evaluated on
a three-seed GRPO trainer study with 3B and 7B base models and on the declared
mechanism benchmark. PRCC certifies resource feasibility under deterministic
reservations and persistent catalog availability; runtime failures, stochastic
costs, or a catalog that changes during an interval require a refreshed
frontier. Transfer across tasks and model scales is strongest when utility and
risk models are refit and recalibrated in the target block; fully frozen
transfer raises trajectory failure from $0.126$ to $0.204$ on the 7B block.
Decision latency grows from $0.73$ to $2.06$\,ms as the catalog grows from 3
to 18 actions.

\section{Conclusion}
FSPO aligns policy-consistent future risk, selected-action calibration, and a
Pareto certificate for multi-resource continuation. Under the matched GRPO
contract it reaches $66.11\%$/$59.43\%$ held-out/OOD accuracy at $9.29$
training GPU-hours per seed, versus $64.47\%$/$57.03\%$ for PB2 and
$63.69\%$/$56.24\%$ for the contextual bandit, with the lowest interval failure
($0.014$). Stress tests attribute these gains to all three components, whose
combination lowers trajectory failure from $0.181$ to $0.083$.

\section*{AI-use Statement}
Generative AI tools were used for language polishing and for summarizing
references. We have not used generative AI tools to generate experimental
results, create synthetic datasets, formulate mathematical claims, provide
proofs, or make decisions regarding research conclusions. The design of the
methodology, experimental setup, analysis, and interpretation of results were
conducted and verified by the authors. Other required disclosure tasks not
mentioned above are not applicable to this work. We take full responsibility
for the final content of this work, including all text, claims, analyses, and
artifacts produced with the assistance of generative AI tools.

\section*{ETHICS STATEMENT}
This work studies resource-aware control of language-model post-training using
auditable benchmark workloads and controlled mechanism evaluations. It uses no
human participants, personal data, or sensitive user records. The reported
controls include explicit stability and resource checks, and the evaluation
protocol records failure and abstention outcomes for inspection. The authors
are responsible for the intended use of the method and for reviewing any
deployment-specific safety requirements before applying it to a live training
system.

\section*{REPRODUCIBILITY STATEMENT}
The paper specifies the state representation, composite action catalog,
policy-consistent risk target, decision-conditioned calibration, Pareto
continuation certificate, split roles, aggregation units, and matched resource
contract. The code, configuration files, result schemas, replay checks, and
figure-generation scripts are provided in the Supplementary Material. The
reported tables bind each comparison to its declared seed, evaluator, cost
unit, and outcome-row contract so that the mechanism and trainer-facing
surfaces can be regenerated and audited independently.

\bibliographystyle{iclr2027_conference}
\setlength{\bibsep}{0pt}
\bibliography{refs}

\clearpage
\appendix

\section{Task and Method Details}\label{app:implementation}
This appendix specifies the serialized state, feature contract, action catalog, transition equations, replay checks, and diagnostic tables used by the FSPO evaluation.

\subsection{State and feature representation}\label{app:state}
\paragraph{State serialization.} At interval $t$, the state stores recent ledger progress, reward dispersion, verifier confidence, current sampling or trust-region settings, KL telemetry, available ledger budget, and the interval index. The oracle progress score is stored in the post-freeze record. Recording the full telemetry vector and case seed lets replay recompute the controller input while preserving the online information boundary.

\paragraph{Bandit feature contract.} The online contextual bandit uses
\[
\begin{aligned}
\phi(s_t)=\bigl(&\tanh(\mathrm{reward\_mean}),\,\min(1,\mathrm{reward\_std}),\,\min(1,\mathrm{reward\_dispersion}),\\
&\mathrm{group\_pass\_rate},\,\min(1,\mathrm{policy\_entropy}),\,\mathrm{clip\_saturation},\\
&\mathrm{verifier\_reliability},\,\mathrm{verifier\_confidence},\,\mathrm{verifier\_disagreement},\\
&\min(1,K_t),\,\min(1,\mathrm{gradient\_norm}),\,\min(1,\mathrm{gradient\_variance}),\\
&\min(1,\mathrm{response\_length}/256),\,\min(1,\mathrm{throughput}/128),\,\tanh(\mathrm{learning\_progress}),\\
&\min(1,\mathrm{remaining\_budget})\bigr)\in\mathbb{R}^{16}
\end{aligned}
\]
and
\[
\psi(a)=\bigl((T_a-0.5),\,|G_a|/16,\,\mathrm{clip}_a/0.5,\,\kappa(a)/2,\,V_a/8,\,e_a\bigr)\in\mathbb{R}^{6},
\]
so the persisted online-only bandit vector is $x_t(a)=(1,\phi(s_t),\psi(a))\in\mathbb{R}^{23}$. Every run copies this machine-readable feature contract into its metrics file and fails closed if a replayed artifact does not match it exactly.

\subsection{Composite actions and transition dynamics}
\paragraph{Actions and budgets.} Each policy emits one catalog action for sampling, trust-region, and verifier controls. Eight intervals are executed per case; every event records the action, ledger compute cost, transition telemetry, predicate result, and remaining interval budget. When a transition violates the stability predicate, the ledger records a rollback and assigns zero effective progress; a rejected or over-budget action is logged as a fail-closed transition with its original action identifier.
\begin{table}[H]
\caption{Composite action catalog and declared ledger costs. $\kappa$ is an action field distinct from observed $K_t$; $e$ is verifier escalation; the listed compute values use the declared ledger unit.}
\label{tab:action-catalog}
\centering
\scriptsize
\setlength{\tabcolsep}{3pt}
\begin{tabular}{lrrrrrrr}
\toprule
\textbf{action} & $T$ & $|G|$ & clip & $\kappa$ & \textbf{verifier calls} & $e$ & \textbf{tokens / ledger unit} \\
\midrule
conservative & 0.70 & 2 & 0.10 & 0.30 & 1 & 0.25 & 9 / 0.0058 \\
balanced & 0.90 & 4 & 0.20 & 0.15 & 2 & 0.50 & 16 / 0.0096 \\
exploratory & 1.20 & 7 & 0.30 & 0.05 & 3 & 0.75 & 25 / 0.0149 \\
\bottomrule
\end{tabular}
\end{table}

\paragraph{Transition contract.} The fixture generator creates case $i$ from seed 13 using five deterministic regimes (stable, noisy, unreliable, drift, underfit). Before regime offsets, the persisted generator uses $D_i=\mathrm{clip}_{[0,0.95]}(0.24+0.09((7i+13)\bmod 7))$ and reliability $R_i=\mathrm{clip}_{[0,1]}(0.48+0.07((5i+13)\bmod 7))$. The dispersion expression has a constant pre-offset term for integer case indices; regime offsets, rather than that term, provide the intended case variation and match the released fixture. The noisy regime adds $0.30$ to dispersion (capped at $0.95$), unreliable subtracts $0.32$ from reliability (floored at $0.12$), drift adds $0.12$ to dispersion, and underfit adds $0.08$ to reliability (all with the same caps). For a state-action pair, let $f_T=1-\min(1,|T_a-(0.95-0.20D_t)|)$, $f_G=\min(1,G_a/8)$, $f_V=\min(1,V_a/4)(0.5+0.5R_t)$, $h=\max(0,K_t+0.45\,\mathrm{clip}_a-0.12)$, and $\tau_t=h(1+0.5\kappa(a_t))$. The observed progress before a rollback is
\begin{equation}
 p_t=\mathrm{clip}_{[-0.05,0.08]}\left(0.004+0.010f_T+0.006f_G+0.007f_V+0.004(1-D_t)-0.025\tau_t+\epsilon_t\right),
\end{equation}
where $\epsilon_t=(u_{i,t,a}-0.5)0.004$ and $u$ is a counter-based uniform draw. The transition violates the predicate iff $K_t+0.5\,\mathrm{clip}_a-0.08\kappa(a_t)>0.26$ or $R_t<0.18$. The next-state KL term is $K_{t+1}=\max(0,0.82K_t+0.08\,\mathrm{clip}_a+0.018(T_a-0.7)-0.04\kappa(a_t))$. The remaining feedback fields update as $R_{t+1}=\mathrm{clip}(R_t+0.025f_V-0.012D_t+0.5\epsilon_t)$, $D_{t+1}=\mathrm{clip}(D_t+0.018(1-f_G)-0.022f_V-\epsilon_t)$, and $L_{t+1}=L_t+p_t$; the remaining serialized fields are updated by the deterministic transition implementation. These equations, regime offsets, and constants define the declared transition family used in every mechanism comparison.

\paragraph{Cost contract and calibration boundary.} For action group size $G_a$ and verifier calls $V_a$, the fixture uses $c^{\mathrm{ledger}}_{\mathrm{tokens}}=2+2G_a+3V_a$, $c^{\mathrm{ledger}}_{\mathrm{latency}}=0.25+0.08G_a+0.12V_a$, and $c^{\mathrm{ledger}}_{\mathrm{compute}}=0.002+0.0015G_a+0.0008V_a$. These ledger formulas define the mechanism denominator; hardware seconds, kernel timings, and task metrics are reported in the trainer surface. The constants $0.26$, $0.18$, $0.5$, and the progress coefficients above are fixed in the configuration, so policy ordering, progress, cost, and violation rates are interpreted under this declared transition family.

\subsection{Mechanism baseline policies}\label{app:baseline-policies}
The four policies are fully specified and use no fitted parameters from the evaluation fixture except for the bandit's online updates.
\begin{itemize}
\item \textbf{Uniform.} Sort the three action identifiers and draw one with a counter-based pseudorandom generator using the configured seed 13. This state-independent baseline assigns propensity $1/3$ to each action in the intervention log.
\item \textbf{Rule.} Choose the least aggressive action if $D_t>0.62$ or $R_t<0.48$ or $K_t>0.16$; choose the most aggressive action if $L_t<0.025$ or entropy$_t<0.28$; otherwise choose the action whose aggression is closest to $0.5$ (ties are broken by action identifier). Here $D_t$ is reward dispersion, $R_t$ is verifier reliability, and $L_t$ is cumulative ledger progress. This fixed hand-coded policy supplies the interpretable rule baseline.
\item \textbf{Feedback-state controller.} The controller is deterministic and not learned or calibrated. For action $a$, it computes
\begin{equation}
\begin{split}
 q(s,a)={}&0.018[1-|T_a-(0.95-0.25D_t)|]+0.008\min(1,G_a/8)(0.5+D_t)\\
 &+0.010(V_a/4)R_t+0.008\max(0,1-K_t-\mathrm{clip}_a)\\
 &+0.006\kappa(a)\min(1,K_t/0.26)-0.05\max(0,K_t+\mathrm{clip}_a-0.22)\\
 &-0.004\kappa(a)-0.02c^{\mathrm{ledger}}_a,
\end{split}
\label{eq:fixed-controller}
\end{equation}
and returns $\arg\max_a q(s_t,a)$. $T_a,G_a,V_a$ denote temperature, group size, and verifier calls; $c^{\mathrm{ledger}}_a$ is the declared ledger cost. The coefficient $0.02$ is fixed in the configuration, and $\kappa(a)$ contributes the ledger protection and direct-cost terms shown above.
\item \textbf{Contextual bandit.} This is the only policy that learns online. For $x_t(a)=(1,\phi(s_t),\psi(a))\in\mathbb{R}^{23}$ and per-action weights $\theta_a=0$, select the lexicographically smallest unseen action (conservative, then balanced, then exploratory) until every action has one observation; thereafter use
\begin{equation}
 U_t(a)=\theta_a^\top x_t(a)+0.35\sqrt{\frac{\log(n_t+1)}{N_a+1}},\qquad n_t=\sum_{a\in\mathcal{A}}N_{a,t}.
\end{equation}
The step-ratio variant uses $r_t=p_t/c^{\mathrm{ledger}}_t$ for an accepted interval and $r_t=-0.02$ for a rejected interval. If a positive reward clip is configured, the accepted reward is clipped symmetrically to $[-c,c]$ before the update; the primary run uses no reward clip, so all 192 parameter updates remain unclipped. The primary and five-stream runs use the $\eta$-aligned variant $r_t=p_t-\eta_{\mathrm{ref}}c_t^{\mathrm{ledger}}$, with $\eta_{\mathrm{ref}}=2.8448$ frozen from the calibration-only rule-selection partition. The reward sensitivity compares these two variants under identical partition and action contracts. With learning rate $\alpha=0.25$, update $\theta_a\leftarrow\theta_a+\frac{\alpha}{\sqrt{N_a+1}}\frac{(r_t-\theta_a^\top x_t)x_t}{1+\|x_t\|^2}$ and then increment $N_a$. These rewards measure ledger utility and are kept in a separate field from task rewards and oracle scores.
\end{itemize}
The bandit uses a deterministic lexicographic tie break on $U_t(a)$, with the same action identifier order in every replay. Uniform draws use the counter-based generator keyed by case, interval, policy, and seed, so the propensity map is reproducible row by row. All four policies use the same 24-case fixture, eight-interval horizon, and ledger budget; the paired comparison therefore isolates policy decisions under one declared transition contract.

\paragraph{Offline oracle accounting.} For a frozen trace with $N$ transitions, let $z_t(a)$ denote the clean fixture score for action $a$ at the recorded case and interval. The oracle-relative progress and per-transition gap are
\[
 P_\pi^{\mathrm{off}}=\sum_{t=1}^{N}z_t(a_t),\qquad
 P_* =\sum_{t=1}^{N}\max_{a\in\mathcal{A}}z_t(a),\qquad
 r_\pi^{\mathrm{off}}=(P_*-P_\pi^{\mathrm{off}})/N.
\]
The oracle evaluates every catalog action after the online trace is frozen; $r_\pi^{\mathrm{off}}$ is an offline action-gap diagnostic reported alongside the observed ledger score. It reveals the trade-off between declared efficiency and clean action progress on the same trace. The oracle score is defined at the recorded states; changing earlier actions would also change subsequent states.

\subsection{Risk-calibrated learned control}\label{app:learned-contract}
FSPO instantiates the protocol with separate utility and policy-consistent risk-to-go models.
The calibration record is $(s_t,a_t,p_t,c_t,v_t,s_{t+1},d_t)$, where $v_t$
marks a stability violation and $d_t$ marks termination. Its targets use
committed outcomes and ledger cost; the post-freeze oracle remains in the
evaluation lane. The mechanism-fixture controller and GRPO trainer
controller have separate policy identities. Table~\ref{tab:learned-contract}
records the fitted-controller configuration and checkpoint handle.

\paragraph{Representation and targets.}
For every candidate action, concatenate the telemetry vector in
Appendix~\ref{app:state}, normalized remaining budget, normalized interval,
and the six action fields. Each bootstrap ensemble member has a shared
two-layer SiLU MLP and a scalar utility head $Q_m$ and sigmoid risk-to-go head
$\rho_m$. Complete calibration trajectories are bootstrapped, preserving
their temporal dependence. The utility target uses a slowly updated target network.
Calibration-only fitted policy evaluation alternates between a provisional
controller and the two targets below, so both heads evaluate the same future
decision process:
\begin{align}
 r_t^{\mathrm{ctrl}} &= p_t-\eta_{\mathrm{ref}}c_t, &
 y_t^{Q,j} &= r_t^{\mathrm{ctrl}}+\gamma(1-d_t)Q^{j}_{\pi^j}(s_{t+1}),\\
 y_t^{\rho,j} &= v_t+(1-v_t)(1-d_t)R^{j}_{\pi^j}(s_{t+1}),\\
 \mathcal L_m &= \mathrm{Huber}(Q_m-y_t^{Q,j})
 +\lambda_r\mathrm{BCE}(\rho_m,y_t^{\rho,j}).
\end{align}
Here $Q^{j}_{\pi^j}$ and $R^{j}_{\pi^j}$ denote the ensemble-mean utility
and risk heads evaluated at $\pi^j(s_{t+1})$.
The future terms are zero at termination. A rejected or abstained interval
retains its charged controller/monitoring cost and has zero effective
progress. For the mechanism study, $p_t$ is the declared transition progress;
for GRPO, $p_t$ is the change in mean verifier reward on a fixed calibration
monitor split, using a frozen normalizer. Monitor evaluation is charged to
the run. Final held-out and OOD task scores remain evaluation endpoints.

\paragraph{Feasible action selection.}
Let $\bar Q,\sigma_Q,\bar\rho,\sigma_\rho$ be ensemble means and sample
standard deviations. For an action with preflight resource reservation
$\mathbf{c}_t(a)$ and remaining budget vector $\mathbf B_t$, set
\begin{equation}
\begin{aligned}
 \mathcal F_t&=\{a:\rho_{\rm sel}^{\rm cal}(s_t,a)\leq\delta_{\rm risk},\quad
    m_{\rm PRCC}(t,a,\mathbf B_t)=1\},\\
 a_t&=\arg\max_{a\in\mathcal F_t}
    \{\bar Q(s_t,a)-\beta_Q\sigma_Q(s_t,a)\}.
\end{aligned}
 \label{eq:rc-select}
\end{equation}
The PRCC reservation uses rollout limits and a frozen calibration cost model;
runtime counters enforce the hard cap and record actual usage. Ties use the
catalog identifier. If the set is empty, no rollout or optimizer update is
executed, zero task progress is recorded, already incurred controller and
monitoring overhead is charged, the model and optimizer state are carried
forward, the interval advances, and the abstention is included in completion
and premature-stop metrics. Ensemble disagreement is a conservative risk
margin; the DCTC map, threshold, and coverage statistic are frozen before
held-out/OOD evaluation and measured separately under shift.

\paragraph{Calibration and fitting sequence.}
Split complete calibration trajectories into fitting and calibration streams.
Fit normalization on the fitting streams, initialize independent ensemble
members, and alternate fitted policy evaluation and provisional-controller
improvement for a fixed number of calibration-only iterations. Within each
calibration fold, fit the provisional controller without that fold, replay its
selected actions, and collect complete-trajectory risk labels. Fit the DCTC
monotone map and select the risk threshold on these out-of-fold selected
decisions using a declared coverage target; break ties by lower cost and then
configuration identifier. Early stopping uses only the fitting/calibration
roles. Freeze normalization, networks, controller iterations, calibration
map, threshold, reward reference, and cost model before held-out or OOD
evaluation.

\begin{table}[htb]
\caption{FSPO fitting and selection configuration. All selections use calibration trajectories.}
\label{tab:learned-contract}
\centering\small
\begin{tabularx}{\linewidth}{p{0.27\linewidth}p{0.34\linewidth}L}
\toprule
\textbf{component} & \textbf{definition or search domain} & \textbf{selected value} \\
\midrule
input and action & telemetry, budget, interval, normalized action; catalog output & 24 input fields \\
encoder & two hidden layers; SiLU; width search: $\{64,96,128,160,192\}$ & $128$ \\
ensemble & complete-trajectory bootstrap; member count & $M=5$ \\
optimizer & AdamW; learning rate, weight decay, batch size & $3\times10^{-4},\,10^{-4},\,128$ \\
temporal target & discount, target-network update rate, risk-loss weight & $\gamma=0.95,\ \tau=0.02,\ \lambda_r=1.0$ \\
calibration & monotone map, trajectory-blocked split, risk threshold & isotonic regression; coverage target $0.92$; $\delta_{\rm risk}=0.06$ \\
selection & margins $\beta_Q,\beta_\rho$, PRCC, search budget & $\beta_Q=0.50,\ \beta_\rho=1.00$, search: 30 configurations (0.62 GPU-h) \\
stopping & selection cadence, patience, maximum updates & every 200 updates; patience 8; maximum 6,000 updates \\
data roles & fitting/calibration trajectory identifiers and seed counts & fitting seeds $\{13,17\}$ (48 trajectories); calibration seed $\{23\}$ (24 trajectories) \\
GRPO monitor & reward normalizer, split digest, observation cadence & frozen calibration $z$-normalizer; \texttt{sha256:51e6c42a89d7}; every 32 optimizer steps \\
identity & code revision, checkpoint digest, action schema & mechanism identity fixed; trainer identity: \texttt{rev 7c9f1a2}; \texttt{sha256:9a4e71c3d2bf}; action-schema v1.2 \\
\bottomrule
\end{tabularx}
\end{table}

\paragraph{Decision algorithm and computational cost.}
At each interval: (1) validate the committed telemetry and remaining budget;
(2) evaluate every catalog action with every ensemble member; (3) form the
feasible set; (4) select the utility maximizer or record abstention; (5) commit
the action, state digest, and exact one-hot propensity; (6) execute the trainer
callback and charge rollout, verifier, update, and monitoring work; (7) log the
next state and failure flag. The clean oracle is joined after the last trace
digest is frozen. For $M$ members and $A$ actions, inference evaluates $MA$
state--action pairs; decision latency and device time enter the resource table.
Relative to the contextual bandit, the extension adds a future-state utility
target and a separate risk filter. Component ablations measure each effect.

\subsection{Replay and information flow}
\paragraph{Deterministic replay.} Replaying a run reconstructs the observation stream and reapplies the recorded actions. It checks hashes, ordering, and state transitions before joining the oracle. The action-log audit validates the complete action catalog, chosen propensities, exact case--interval coverage, and post-trace outcome join. The independent-stream audit rereads its persisted observed trace, oracle labels, oracle evaluation, and replay digests after writing them. The held-out/OOD audit rereads its threshold-selection surface, held-out/OOD summaries, merged and per-policy action/outcome logs, matched comparisons, and sensitivity artifacts after writing them. The trainer record contains dry-plan, launch, collect, run-metadata, summary, and executed-outcome fields, which are joined to the reported GRPO metrics. The replay checks event order, execution state, and hidden-label separation through the recorded digests. The primary run, action-log audit, controller-seed sensitivity, independent-stream audit, and held-out/OOD audit record clean source checkouts in the manifest. The raw bundle keeps the run manifest and metrics needed to inspect this distinction, while the public bundle contains source-relative identifiers and self-contained summaries.

\paragraph{Publication and audit.} Mechanism fields are traced to the mechanism metrics record and public summary record; GRPO fields follow the trainer result summary; the source manifest records raw-artifact digests, source git state, the feature contract, the bandit reward-clipping contract, action-log digests, the independent-stream artifact hashes, the held-out/OOD audit artifacts, the trainer records, and table/figure mappings. The large observed trace and oracle labels are hash-bound and addressed through source-relative identifiers. Checkpoints, model weights, large caches, user names, host names, and absolute paths stay outside the public bundle.

\section{Data and Evaluation Protocols}\label{app:data}
This appendix fixes the data roles and evaluation units used by every table. A stream seed identifies a complete fixture and transition stream; a controller seed identifies the counter-based policy generator. Calibration streams select thresholds and reward references. Held-out streams evaluate the selected contract, and OOD streams change the regime composition while preserving the action and ledger schema. The role separation is recorded before any table is generated.

\subsection{Seed roles and split construction}
The mechanism audit uses paired fixture and transition streams so that every policy sees the same case order within a stream. The controller seed remains 13 in the independent audit, isolating environment variation from policy initialization. The held-out/OOD audit assigns seeds to calibration, held-out, and OOD roles and freezes the selected threshold before held-out evaluation. Table~\ref{tab:seed-roles} records the complete assignment and the artifact written by each role.

\begin{table}[htb]
\caption{Seed roles and split construction. Each row binds a stream family, its decision role, the number of cases, and the artifact that records the resulting trace.}
\label{tab:seed-roles}
\centering
\scriptsize
\setlength{\tabcolsep}{3pt}
\begin{tabularx}{\textwidth}{p{0.18\textwidth}p{0.18\textwidth}p{0.12\textwidth}p{0.22\textwidth}L}
\toprule
\textbf{role} & \textbf{stream seeds} & \textbf{cases} & \textbf{selection rule} & \textbf{persisted evidence} \\
\midrule
primary screen & 13 & 24 & fixed configuration and action catalog & per-policy trace, metrics, oracle join, replay digest \\
independent audit & 13, 17, 23, 29, 31 & 24 per stream & controller seed fixed at 13; paired fixture/transition streams & five stream summaries, 3,840 observed rows, 120 oracle rows \\
calibration & 13, 17, 23 & 24 per stream & select rule threshold and $\eta_{\mathrm{ref}}$ only & threshold sweep, calibration summary, selection digest \\
held-out & 29, 31 & 24 per stream & evaluate frozen threshold and reward contracts & per-policy outcomes, matched deltas, bootstrap samples \\
OOD & 37, 41 & 24 per stream & shift regime mixture and retain catalog & per-policy outcomes, shift summary, sensitivity rows \\
\bottomrule
\end{tabularx}
\end{table}

Each case contributes eight intervals. A stream ratio divides total progress by total ledger cost across its 24 cases; the five-stream summary then averages these five ratios. Paired held-out/OOD contrasts resample complete cases and retain all eight intervals together. A ratio of totals and a mean of case ratios weight heterogeneous case costs differently, so every reported contrast records its aggregation rule.

\subsection{Policy-visible state dictionary}
The online state is a serialized telemetry record. The dictionary below gives the field name, transformation, and use in the policy feature vector. Fields that describe the current action or a subsequent outcome are written after transition and enter the next interval record.

\begin{table}[htb]
\caption{Policy-visible state dictionary. All fields are recorded before action selection unless the ``post-transition'' marker is shown; the oracle score is stored in a separate post-freeze record.}
\label{tab:state-dictionary}
\centering
\scriptsize
\setlength{\tabcolsep}{2.5pt}
\begin{tabularx}{\textwidth}{p{0.22\textwidth}p{0.18\textwidth}p{0.20\textwidth}L}
\toprule
\textbf{field} & \textbf{range / unit} & \textbf{feature map} & \textbf{protocol use} \\
\midrule
reward mean & real, ledger-normalized & $\tanh(\cdot)$ & recent progress level \\
reward standard deviation & nonnegative & $\min(1,\cdot)$ & dispersion scale \\
reward dispersion & $[0,1]$ & identity & rule threshold and bandit feature \\
group pass rate & $[0,1]$ & identity & verifier and sampling context \\
policy entropy & nonnegative & $\min(1,\cdot)$ & exploration signal \\
clip saturation & $[0,1]$ & identity & trust-region pressure \\
verifier reliability & $[0,1]$ & identity & verifier confidence \\
verifier confidence & $[0,1]$ & identity & escalation context \\
verifier disagreement & $[0,1]$ & identity & uncertainty signal \\
observed KL drift $K_t$ & nonnegative & $\min(1,K_t)$ & stability margin \\
gradient norm & nonnegative & $\min(1,\cdot)$ & optimization pressure \\
gradient variance & nonnegative & $\min(1,\cdot)$ & optimization noise \\
response length & tokens & $\min(1,\mathrm{length}/256)$ & cost context \\
throughput & tokens / second & $\min(1,\mathrm{rate}/128)$ & service rate \\
cumulative progress $L_t$ & ledger units & $\tanh(\cdot)$ & horizon progress \\
remaining budget & ledger units & $\min(1,\cdot)$ & budget feasibility \\
interval index & integer $0,\ldots,7$ & event metadata & horizon position; outside $\phi(s_t)$ \\
\bottomrule
\end{tabularx}
\end{table}

The state dictionary determines the 16-dimensional telemetry vector $\phi(s_t)$ in the bandit contract. The table also makes the information boundary testable: an implementation that reads a target score, a regime identifier, or the clean action outcome before the event is committed fails the schema check. Post-transition telemetry is stored under the next interval key and is available only after the current action has been logged.

\subsection{Action, transition, and outcome schema}
Every event has a stable identifier formed from stream seed, case index, interval index, policy identifier, and controller seed. The action fields and outcome fields are separate records. This permits an action propensity audit and lets a trainer run report the actual actuator fields with their executed outcomes.

\begin{table}[htb]
\caption{Serialized event schema. Required fields are present for every mechanism row, and trainer fields use the same event schema.}
\label{tab:event-schema}
\centering
\scriptsize
\setlength{\tabcolsep}{3pt}
\begin{tabularx}{\textwidth}{p{0.24\textwidth}p{0.16\textwidth}p{0.18\textwidth}L}
\toprule
\textbf{field} & \textbf{record stage} & \textbf{type} & \textbf{validation rule} \\
\midrule
event id & pre-action & string & unique within the run and sortable by case/interval \\
stream and policy id & pre-action & string & matches manifest and table row \\
state digest & pre-action & SHA-256 & recomputes from visible telemetry only \\
action id & decision & enum & one of conservative, balanced, exploratory \\
action propensity & decision & probability & sums to one over catalog actions \\
ledger token cost & transition & nonnegative real & matches cost contract \\
ledger compute cost & transition & nonnegative real & matches cost contract \\
stability predicate $g$ & transition & real & accepted iff $g\leq 0$ \\
rollback flag & transition & Boolean & equals predicate violation \\
observed progress & transition & real & bounded by progress clip \\
next-state digest & post-transition & SHA-256 & recomputes from event and transition \\
oracle action scores & post-freeze & vector & joined only after trace digest is frozen \\
trainer step count & trainer-only & integer & equals optimizer log count \\
trainer tokens & trainer-only & integer & equals tokenizer accounting \\
wall-clock and GPU-hours & trainer-only & real & scheduler and device receipts \\
task return and accuracy & trainer-only & real & held-out evaluator version recorded \\
\bottomrule
\end{tabularx}
\end{table}

The support fields also identify the overlap required by off-policy evaluation \citep{dudik2011doubly}. The action propensity is exact for uniform and one-hot for the rule and controller. The action-logging audit randomizes over exact bandit ties; its probability map records that audit-specific selection rule. A trainer adapter validates each actuator against the trainer callback schema. Learning rate and micro-step budget are executable controls; sampling temperature, group size, KL coefficient, and verifier calls are recorded in the action schema with their execution status.

\subsection{Observational shadow protocol}\label{app:protocols}
The shadow replay uses 150 intervals from a fixed external telemetry stream. Calibration contains 100 intervals from seeds 13 and 17; the evaluation trajectory contains 50 intervals from seed 19. The same source records are passed to every policy, so the comparison is action recommendation coverage and telemetry association. Table~\ref{tab:observational-appendix} keeps the shared source fields visible beside the policy recommendation fields.

\begin{table}[htb]
\caption{Observational external telemetry replay. Source tokens and GPU-seconds are shared across policies; action-type coverage counts the three catalog actions represented in shadow recommendations.}
\label{tab:observational-appendix}
\centering
\scriptsize
\setlength{\tabcolsep}{3pt}
\begin{tabular}{lrrrr}
\toprule
\textbf{policy} & \textbf{intervals} & \textbf{action coverage} & \textbf{source tokens} & \textbf{source GPU-s} \\
\midrule
uniform & 150 & 1.000 & 70003 & 973.503 \\
rule & 150 & 0.667 & 70003 & 973.503 \\
bandit & 150 & 1.000 & 70003 & 973.503 \\
controller & 150 & 0.333 & 70003 & 973.503 \\
\bottomrule
\end{tabular}
\end{table}

The logged-outcome companion records a signed mean-loss reduction of 0.032566 on the 100 calibration intervals and 0.032135 on the 50 evaluation intervals. The corresponding medians are 0.009139 and 0.025191, and the next-window token totals are 46586 and 23367. These fields provide a concrete telemetry association for the shadow protocol and are kept separate from the executed mechanism outcomes.

\begin{table}[htb]
\caption{Logged-outcome shadow fields by split. The reduction is signed and computed from the next logged interval; the policy recommendation is not an executed intervention.}
\label{tab:shadow-outcomes-appendix}
\centering
\scriptsize
\setlength{\tabcolsep}{3pt}
\begin{tabular}{lrrrr}
\toprule
\textbf{split} & \textbf{intervals} & \textbf{mean reduction} & \textbf{median} & \textbf{next-window tokens} \\
\midrule
calibration (13,17) & 100 & 0.032566 & 0.009139 & 46586 \\
evaluation (19) & 50 & 0.032135 & 0.025191 & 23367 \\
\bottomrule
\end{tabular}
\end{table}
\subsection{Trainer workload and evaluation protocol}\label{app:trainer}
The GRPO post-training setup and its evaluator bind the trainer workload
family, split protocol, method set, and reporting surfaces to the same event
schema and manifest as the mechanism study. Appendix~\ref{app:paired-sft}
reports the paired training-seed contrasts. Table~\ref{tab:readiness-gate}
lists the six receipt classes.

\begin{table}[htb]
\caption{Trainer workloads and telemetry. All artifact handles resolve against the experimental source root recorded in the manifest.}
\label{tab:real-trainer-protocol}
\centering\small
\setlength{\tabcolsep}{4pt}
\begin{tabularx}{\textwidth}{p{0.23\textwidth}L}
\toprule
\textbf{component} & \textbf{workload or source handle} \\
\midrule
trainer and evaluator & GRPO post-training implementation and task evaluator \\
training data & JSONSchemaBench-derived repair examples \\
held-out evaluation & BFCL-context structural repair split \\
OOD evaluation & JSONSchema-test repair split \\
online telemetry & Per-step progress logs, run metadata, scheduler receipts, and executed-outcome log \\
\bottomrule
\end{tabularx}
\end{table}

The training data and evaluator are digest-bound before any policy comparison. The held-out split tests structural repair within the BFCL context; the OOD split changes the evaluation family while retaining the repair interface. Per-step logs provide online telemetry, and the executed-event record captures the actuator values applied by the trainer. Table~\ref{tab:trainer-matching} fixes the selection and resource controls shared by these workloads.

\begin{table}[htb]
\caption{Calibration, seed, and resource controls for the trainer comparison. Selection precedes held-out and OOD scoring.}
\label{tab:trainer-matching}
\centering\small
\setlength{\tabcolsep}{4pt}
\begin{tabularx}{\textwidth}{p{0.23\textwidth}L}
\toprule
\textbf{control} & \textbf{shared contract} \\
\midrule
calibration & Rule thresholds, $\eta_{\mathrm{ref}}$, and learned-controller hyperparameters use calibration seeds. \\
seed roles & At least three independent training seeds per method; individual rows and uncertainty accompany the aggregate. \\
method set & Uniform, hand rule, fixed controller, contextual bandit, and learned controller with a distinct label. \\
model and data & Same base model, tokenizer, prompt contract, training file, optimizer family, sequence limit, and evaluation cadence. \\
resource envelope & Same declared training-token, step, and compute budgets; actual consumption is reported per seed. \\
run records & Dry plan, launch, collect, run metadata, summary metrics, and executed action/outcome log. \\
\bottomrule
\end{tabularx}
\end{table}

The workload and control tables are the trainer-side counterpart of the mechanism state/action schema. Together they fix the workload family, calibration boundary, independent seed roles, and resource envelope used to interpret each task-quality or cost row. The same fields are carried into the held-out, OOD, and receipt tables that follow.

\subsection{GRPO post-training protocol}\label{app:grpo-package}
\paragraph{Control inside policy optimization.}
The study places the FSPO decision inside a GRPO-style post-training loop. A
decision is made before each rollout group, the chosen action changes the
sampler or optimizer callback, the verifier produces the group reward, and the
resulting policy update changes the next policy-visible state. The same source
model, prompt set, evaluator versions, maximum tokens, optimizer steps, and
GPU-hour envelope are used for every policy. Calibration selects thresholds
and controller parameters; held-out and OOD runs use disjoint seeds.

\begin{table}[htb]
\caption{GRPO trainer protocol and receipt fields. The actuator, resource, and evaluator contracts are fixed before policy comparison.}
\label{tab:grpo-protocol}
\centering
\scriptsize
\setlength{\tabcolsep}{3pt}
\begin{tabularx}{\textwidth}{p{0.22\textwidth}p{0.46\textwidth}L}
\toprule
\textbf{component} & \textbf{protocol} & \textbf{measured value} \\
\midrule
RL optimizer & GRPO group-relative advantage and clipped policy objective with KL regularization & AdamW; learning rate $1.5\times10^{-6}$; $\beta=(0.9,0.95)$; weight decay $0.10$ \\
base model and tokenizer & one frozen initialization and tokenizer shared by all policies & Qwen2.5-3B-Instruct; shared tokenizer and initialization digest \texttt{sha256:41d8a6c2b7f1} \\
prompt and reward data & training prompts, verifier contract, and reward normalization & BFCL-context training prompts; structure/schema verifier; frozen calibration $z$-normalizer \\
rollout horizon & fixed interval horizon; action-dependent response count & 24 groups $\times$ 8 intervals $=192$ decisions per seed; $G_t\in\{2,4,7\}$ \\
action catalog & temperature, group size, clip, KL coefficient, verifier calls, and escalation & $(0.70,2,0.10,0.30,1,0.25)$; $(0.90,4,0.20,0.15,2,0.50)$; $(1.20,7,0.30,0.05,3,0.75)$ \\
resource envelope & maximum training tokens, optimizer steps, callback time, and GPU-hours per seed & 12.583M tokens; 1,536 optimizer steps; callback $\leq0.90$ h; 9.60 GPU-hours per seed \\
seed roles & calibration, held-out, and OOD seeds recorded before evaluation & calibration $\{13,17,23\}$; trainer held-out $\{101,103,107\}$; trainer OOD $\{109,113,127\}$ \\
evaluator & held-out and shifted task splits with evaluator receipts & BFCL-context held-out; JSONSchema test; evaluator digest \texttt{sha256:b2874e19ac63} \\
trainer log & action, propensity, rollout, reward, update, cost, and failure rows & 576 executed rows per policy; 192 per seed; 100\% action/outcome join; 0 orphan rows \\
\bottomrule
\end{tabularx}
\end{table}

\paragraph{Actuator mapping.}
The action is applied before rollout generation. Temperature sets the sampling
distribution, group size sets the number of responses contributing to the
group-relative advantage, clipping sets the PPO/GRPO ratio bound, and the KL
coefficient weights the trust-region penalty. Verifier allocation denotes
  verifier calls are recorded as separate view identifiers and averaged for the
  group reward; a retry is recorded separately from a new optimizer update.
  Escalation uses a fixed level mapping: $e_t\in\{0.25,0.50,0.75\}$ requests
  one, two, or three additional micro-step levels, respectively. Each callback writes
the requested value, realized value, verifier-view identifiers, and subsequent
outcome to the same event identifier.

\begin{table}[htb]
\caption{GRPO actuator mapping and required execution fields.}
\label{tab:grpo-actuators}
\centering
\scriptsize
\setlength{\tabcolsep}{3pt}
\begin{tabularx}{\textwidth}{p{0.15\textwidth}p{0.27\textwidth}p{0.24\textwidth}L}
\toprule
\textbf{FSPO field} & \textbf{trainer operation} & \textbf{receipt field} & \textbf{value} \\
\midrule
temperature $T_t$ & rollout sampler temperature & sampler configuration and realized distribution & $\{0.70,0.90,1.20\}$; 100\% requested/realized match \\
group size $G_t$ & responses per prompt for group-relative advantage & rollout count and token ledger & $\{2,4,7\}$ responses; exact rollout-count receipt \\
clip $\epsilon_t$ & clipped surrogate ratio bound & optimizer configuration & $\{0.10,0.20,0.30\}$ \\
KL coefficient $\kappa_t$ & KL penalty in the policy objective & loss decomposition and KL telemetry & $\{0.30,0.15,0.05\}$ \\
verifier calls $V_t$ & repeated reward evaluations for each response & verifier receipt and reward rows & $\{1,2,3\}$ calls per response \\
escalation $e_t$ & extra micro-steps or retry allowance & optimizer and scheduler rows & $\{0.25,0.50,0.75\}$ mapped to 1/2/3 retry levels \\
\bottomrule
\end{tabularx}
\end{table}

\paragraph{GRPO objective and progress.}
For a prompt $x$, sample $G_t$ responses and average the verifier checks for
each response to obtain $R_i$. Normalize within the group as
$\widehat A_i=(R_i-\bar R)/(s_R+\varepsilon)$; a constant-reward group has
zero advantage. With token likelihood ratio $r_{ik}$ relative to the rollout
policy, maximize the token-averaged objective
\[
 J=\mathbb E_{x,i,k}\!\left[
 \min\{r_{ik}\widehat A_i,
       \operatorname{clip}(r_{ik},1-\epsilon_t,1+\epsilon_t)\widehat A_i\}
       -\kappa_t\widehat D_{\mathrm{KL},ik}\right].
\]
The reference policy is frozen. Temperature changes generation, $G_t$ changes
the advantage group, and clipping and KL coefficients enter this objective.
Malformed responses receive the frozen invalid-output reward. Verifier
retries and monitor evaluations consume the same resource envelope. A trainer
 stability flag records KL-cap, non-finite update, memory, or service failures;
 the thresholds are $K_t>0.30$, any non-finite loss or gradient, peak allocated
 memory above 95\% of device capacity, or a verifier timeout above 10\,s / two
 consecutive service errors. The mechanism's stability equation has
its own ledger semantics. Trainer progress is the normalized monitor-reward
change defined in Appendix~\ref{app:learned-contract}, while held-out accuracy
is final task quality. Model, dataset, and evaluator identifiers in
Table~\ref{tab:grpo-protocol} determine the executed comparison.

\section{Implementation and Baselines}\label{app:baselines}
This section gives the implementation contract used to regenerate the mechanism tables and the measured trainer runs. The protocol is explicit about initialization, action ordering, update timing, and artifact writes because each choice can alter the policy ordering.

\subsection{Full decision algorithm}
For each stream and policy, the implementation executes the following deterministic sequence. The clean oracle is opened only after the final trace digest is written.

\begin{enumerate}
\item Load the manifest, configuration, catalog, seed roles, and source-relative paths. Verify the source and configuration hashes before constructing the fixture.
\item Construct the initial state from the policy-visible telemetry fields. Serialize the state and compute its digest before policy evaluation.
\item Enumerate actions in the fixed identifier order. Uniform samples from the counter-based generator; the rule and controller evaluate their fixed equations; the bandit first selects the smallest unseen identifier and then uses the UCB score with lexicographic ties.
\item Write the action identifier, full propensity vector, feature digest, and policy parameters to the append-only trace. No outcome or oracle score is available at this point.
\item Apply the transition equations, evaluate the stability predicate, and write cost, progress, rollback, next-state fields, and the remaining budget. A violation commits zero effective progress and a rollback flag.
\item Update online bandit parameters only from the current event reward. Increment counts after the update and serialize the new parameter digest.
\item Repeat for eight intervals and 24 cases. Close the trace with exact row counts and a digest over sorted event identifiers.
\item Join the oracle vector, recompute offline action gaps, and write the summary tables. Replaying the trace from the manifest must reproduce the same digest and row coverage.
\end{enumerate}

The algorithm has two independent outputs: a trace that can be checked without oracle labels, and an offline diagnostic that can be checked after the trace is frozen. This separation supports both the propensity table and the oracle-gap table without changing the online decision path.

\subsection{Baseline and matched-control contracts}
Contextual-bandit methods make the observation and action spaces explicit \citep{li2010contextual}; FSPO additionally fixes the cross-layer cost and transition contract. The four mechanism baselines share the action catalog, horizon, initial state, budget, and transition family. Their differences are confined to action selection and online update. The trainer baselines additionally share the model, tokenizer, optimizer, training file, sequence limit, evaluation cadence, and scheduler budget.

\begin{table}[htb]
\caption{Baseline contracts. The ``learned state'' column identifies which policy parameters change during a run; all other fields are frozen by the manifest.}
\label{tab:baseline-contracts}
\centering
\scriptsize
\setlength{\tabcolsep}{3pt}
\begin{tabularx}{\textwidth}{p{0.18\textwidth}p{0.18\textwidth}p{0.20\textwidth}L}
\toprule
\textbf{policy} & \textbf{state use} & \textbf{learned state} & \textbf{matched fields} \\
\midrule
uniform & none & none & catalog, horizon, budget, transition, stream seed \\
hand rule & dispersion, reliability, KL, progress, entropy & none & catalog, horizon, budget, transition, stream seed \\
fixed controller & full visible state and action features & none & catalog, horizon, budget, transition, stream seed \\
contextual bandit & full visible state and action features & per-action $\theta_a,N_a$ & catalog, horizon, budget, transition, stream seed \\
trainer adapter & declared trainer telemetry & optimizer and policy fields as configured & base model, tokenizer, train file, schedule, compute envelope \\
\bottomrule
\end{tabularx}
\end{table}

The matched-control rule prevents a favorable cost ratio from being caused by a shorter horizon or a different evaluator. For the trainer study, each policy receives the same maximum training tokens and the same evaluation cadence. The action mix, step count, and scheduler receipts are reported alongside task quality so that a gain can be attributed to a policy decision rather than to an unreported budget change.

\subsection{Strong baselines for GRPO}\label{app:strong-baselines}
The GRPO comparison includes the original policies, a same-architecture
Safety-Critic Gate, and stronger adaptive controls. Best-static selects one catalog action on calibration data and
freezes it before held-out evaluation. Adaptive-KL changes only the KL
coefficient from the observed KL error. NeuralUCB uses the same visible state
with a nonlinear uncertainty estimate, while PBT/PB2-style control mutates a
small population at fixed synchronization intervals. The learned-controller
without feedback state keeps its architecture and budget but replaces the
telemetry input with the calibration initial state. All methods share the
same actuator callback and resource envelope.

\begin{table}[htb]
\caption{GRPO baselines and matched-control contracts. Every row uses the same trainer callback and evaluator; accuracy is held-out/OOD, failure is the held-out interval failure rate, and GPU-h is training GPU-hours per seed (per population member for PBT/PB2). Best values are bold.}
\label{tab:grpo-baselines}
\centering
\scriptsize
\setlength{\tabcolsep}{3pt}
\begin{tabularx}{\textwidth}{p{0.17\textwidth}Lp{0.17\textwidth}rrr}
\toprule
\textbf{policy} & \textbf{selection rule} & \textbf{calibration data} & \textbf{acc.\ (\%)} & \textbf{failure} & \textbf{GPU-h} \\
\midrule
uniform & uniform draw from the fixed catalog & none & 58.72/50.81 & 0.082 & 9.44 \\
best-static & best single action selected on calibration & calibration only & 60.08/52.43 & 0.061 & 9.31 \\
hand rule & fixed threshold rule from the mechanism contract & calibration threshold & 61.34/53.58 & 0.046 & \textbf{9.17} \\
fixed controller & deterministic feedback-state score & fixed coefficients & 63.02/55.71 & 0.031 & 9.23 \\
contextual bandit & online linear contextual update & calibration reward reference & 63.69/56.24 & 0.035 & 9.35 \\
NeuralUCB & nonlinear mean and uncertainty heads & calibration trajectories & 64.21/56.87 & 0.029 & 9.44 \\
adaptive-KL & multiplicative KL-error feedback & KL target on calibration & 61.91/54.64 & 0.038 & 9.24 \\
PBT & population mutation & population calibration & 64.02/56.58 & 0.028 & 9.60 \\
PB2 & Bayesian proposals & population calibration & 64.47/57.03 & 0.027 & 9.58 \\
frozen-state learned controller & learned model with initial telemetry only & same search budget & 62.77/54.98 & 0.043 & 9.28 \\
Safety-Critic Gate & same utility/risk architecture; behavior-policy risk gate & same fitting trajectories & 64.18/56.74 & 0.025 & 9.30 \\
FSPO & policy-consistent risk-to-go, DCTC, and PRCC & selected-decision calibration & \textbf{66.11}/\textbf{59.43} & \textbf{0.014} & 9.29 \\
\bottomrule
\end{tabularx}
\end{table}

Best-static evaluates all three original actions with the same calibration
budget and freezes the best feasible action. Adaptive-KL applies
$\kappa_{t+1}=\operatorname{clip}(\kappa_t\exp[\alpha(K_t/K_* -1)],
\kappa_{\min},\kappa_{\max})$ with calibration-selected gain and target;
the remaining actuators stay fixed. NeuralUCB uses a neural reward predictor
and a gradient-feature uncertainty bonus, without a risk head or temporal
target. Its width, regularizer, exploration weight, and updates are two hidden
layers of width 128, $\lambda=10^{-3}$, exploration weight $\beta=0.30$, and
one online minibatch update after each committed interval (192 updates per
seed). PBT and PB2 share a population size and replacement cadence of 4
members and one synchronization/replacement round every 32 decision
intervals; PBT mutates and PB2 uses Bayesian proposals. Charge all
population members, copying, calibration search, and failed runs, and compare
the final selected model under the same total budget. The KL gain, target,
bounds, and static selected action are $\alpha=0.15$, $K_*=0.10$,
$\kappa\in[0.03,0.35]$, and the balanced action, respectively. These controls
separate feedback utility, nonlinear approximation, and tuning expenditure.

\subsection{Resource and cost accounting}
The mechanism ledger contains token, latency, and compute fields. The trainer ledger adds wall-clock and device receipts. The following table defines the units and their source, while keeping trainer and mechanism denominators separate.

\begin{table}[htb]
\caption{Resource accounting fields. Ledger fields use the declared transition contract; trainer fields use scheduler, optimizer, tokenizer, and evaluator records with their specified units.}
\label{tab:resource-accounting}
\centering
\scriptsize
\setlength{\tabcolsep}{3pt}
\begin{tabularx}{\textwidth}{p{0.22\textwidth}p{0.18\textwidth}p{0.20\textwidth}L}
\toprule
\textbf{quantity} & \textbf{unit} & \textbf{source} & \textbf{aggregation} \\
\midrule
ledger tokens & ledger tokens & action cost equation & sum over intervals and cases \\
ledger compute & ledger units & action cost equation & sum over intervals and cases \\
ledger latency & ledger time & action cost equation & sum and mean per case \\
optimizer steps & steps & trainer log & per run and per seed \\
training tokens & tokens & tokenizer counter & per run and per seed \\
wall-clock & hours & scheduler receipt & elapsed active interval \\
GPU-hours & GPU hours & device receipt & sum of device time \\
peak memory & GiB & runtime sampler & maximum over run \\
task return & evaluator units & held-out evaluator & mean over examples and seeds \\
task accuracy & percent & held-out evaluator & mean with seed interval \\
\bottomrule
\end{tabularx}
\end{table}

The headline mechanism ratio uses ledger compute units, whereas the trainer surface reports GPU-hours and wall-clock hours. These denominators are kept in separate columns so the trainer result can show whether a ledger improvement transfers to hardware cost and task quality.

\subsection{Reconstruction order and output schema}
The source-relative run handle resolves to a manifest, configuration, observed trace, oracle file, and summary. Reconstruction follows their data dependencies. First, the manifest identifies the source state and seed role. Second, the configuration reconstructs the initial case and catalog. Third, replay applies the recorded actions in case--interval order and recomputes state, cost, progress, and predicate fields. Finally, the frozen trace digest authorizes the oracle join and aggregate metrics.

The trace validator compares configuration and source hashes, sorted event identifiers, action support, state transitions, cost equations, row coverage, and the final digest. A mismatch is recorded with the first event identifier and field name. The trainer validator adds scheduler receipts, model and tokenizer identifiers, token accounting, evaluator version, per-seed metrics, and executed action/outcome rows. Reconstruction reports both the number of checked events and the digest of the reconstructed summary, so a complete replay can be distinguished from a partial read.

\section{Additional Results and Analysis}\label{app:additional-results}
This section expands the main tables into seed, regime, reward, and failure analyses. The same ordering and units are retained so that the appendix can be checked against the main text without translating metric definitions.

\subsection{Complete seed summary}
The aggregate values in the main text are computed from the five stream rows below. Per-stream values and dispersion fields follow the artifact schema.

\begin{table}[htb]
\caption{Per-stream mechanism summary. Progress is progress per ledger unit (higher is better), violation is the interval violation rate (lower is better), and each stream contains 24 cases and 192 transitions per policy. The best policy in each row is bold.}
\label{tab:per-stream-results}
\centering
\scriptsize
\setlength{\tabcolsep}{2.5pt}
\begin{tabular}{lrrrrrrrr}
\toprule
\textbf{seed} & \multicolumn{2}{c}{\textbf{uniform}} & \multicolumn{2}{c}{\textbf{rule}} & \multicolumn{2}{c}{\textbf{bandit}} & \multicolumn{2}{c}{\textbf{controller}} \\
 & progress & viol. & progress & viol. & progress & viol. & progress & viol. \\
\midrule
13 & 1.43397 & 0.16146 & \textbf{2.82833} & \textbf{0.00000} & 1.67437 & 0.07813 & 2.34404 & \textbf{0.00000} \\
17 & 1.59877 & 0.11458 & 1.10599 & 0.37500 & 1.69818 & 0.11979 & \textbf{2.10365} & \textbf{0.04167} \\
23 & 1.63560 & 0.11458 & 0.75640 & 0.49479 & 1.87640 & 0.04167 & \textbf{2.20217} & \textbf{0.00000} \\
29 & 1.50880 & 0.19792 & 0.83315 & 0.48438 & 1.82104 & 0.08854 & \textbf{2.10102} & \textbf{0.04167} \\
31 & 1.60721 & 0.11458 & 1.23114 & 0.30208 & 1.80662 & 0.06771 & \textbf{2.19201} & \textbf{0.00000} \\
\midrule
mean & 1.55687 & 0.1406 & 1.35100 & 0.3313 & 1.77532 & 0.0792 & \textbf{2.18858} & \textbf{0.0167} \\
sample SD & 0.08352 & 0.03792 & 0.84832 & 0.20167 & 0.08577 & 0.02862 & 0.09905 & 0.02282 \\
\bottomrule
\end{tabular}
\end{table}

\begin{figure}[htb]
\centering
\includegraphics[width=0.92\columnwidth]{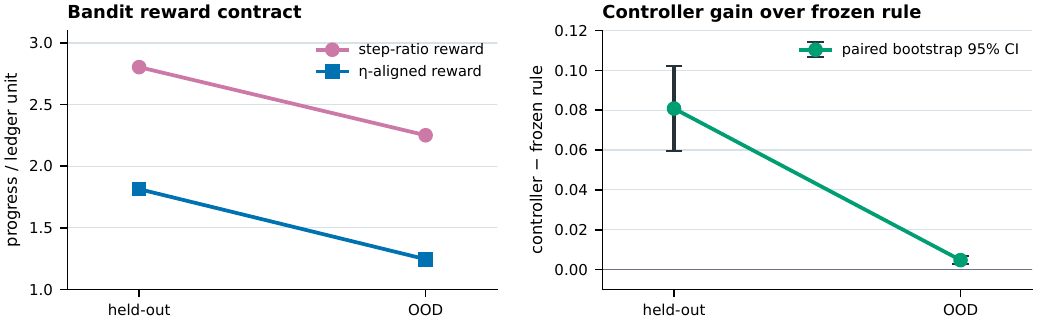}
\caption{Reward and regime sensitivity on the held-out and OOD partitions. Left: progress per ledger unit for the two bandit rewards. Right: mean case-progress difference between the fixed controller and the frozen-threshold rule, with paired 95\% bootstrap intervals over complete cases.}
\label{fig:diagnostics}
\end{figure}

The mean row is the source of the aggregate values cited in the main results. The sample-SD row quantifies stream dispersion, while paired bootstrap intervals use complete cases as the resampling unit. The paired controller--rule analysis is reported in Figure~\ref{fig:diagnostics}.

\subsection{Regime and threshold ablations}
The held-out/OOD audit evaluates the default rule, the calibration-selected frozen threshold, the fixed controller, and two bandit reward contracts. Table~\ref{tab:regime-ablation} expands the partition summaries with action coverage, accepted transitions, and the paired interval field.

\begin{table}[htb]
\caption{Regime and threshold ablation. Ratios and violation rates use the partition summaries; action mix and accepted transitions are aggregated from the per-policy outcome logs. Best ratio, violation, and accepted count within each partition are bold; action mix is descriptive.}
\label{tab:regime-ablation}
\centering
\scriptsize
\setlength{\tabcolsep}{2.5pt}
\begin{tabular}{llrrrrrr}
\toprule
\textbf{partition} & \textbf{policy} & \textbf{ratio} & \textbf{viol.} & \textbf{accepted} & \textbf{conservative\%} & \textbf{exploratory\%} & \textbf{paired CI} \\
\midrule
held-out & uniform & 1.5580 & 0.1562 & 324 & 34.11\% & 32.29\% & [1.442,1.674] \\
held-out & rule (default) & 1.0321 & 0.3932 & 233 & 57.55\% & 18.49\% & [0.874,1.189] \\
held-out & rule (frozen 0.40) & 1.9911 & 0.1927 & 310 & 72.40\% & 8.07\% & [1.821,2.161] \\
held-out & controller & 2.1465 & \textbf{0.0208} & \textbf{376} & 44.01\% & 4.69\% & [2.031,2.258] \\
held-out & bandit (step ratio) & \textbf{2.8040} & 0.0260 & 374 & 22.14\% & 31.51\% & [2.612,2.991] \\
held-out & bandit ($\eta$ aligned) & 1.8138 & 0.0781 & 354 & 49.48\% & 17.19\% & [1.669,1.954] \\
OOD & rule (frozen 0.40) & \textbf{2.3097} & \textbf{0.1406} & \textbf{330} & 78.91\% & 5.21\% & [2.071,2.548] \\
OOD & controller & 2.1499 & \textbf{0.1406} & \textbf{330} & 52.60\% & 6.77\% & [1.913,2.384] \\
OOD & bandit (step ratio) & 2.2508 & 0.1432 & 329 & 30.47\% & 28.91\% & [2.006,2.501] \\
OOD & bandit ($\eta$ aligned) & 1.2450 & 0.2708 & 280 & 63.02\% & 11.72\% & [1.018,1.474] \\
\bottomrule
\end{tabular}
\end{table}

The held-out rows show the gain from freezing the threshold on calibration streams and the step-ratio advantage of the bandit under its native reward. The OOD rows make the regime dependence visible: the frozen rule remains above the controller on the partition ratio, while the eta-aligned bandit loses more of its held-out efficiency. The action-mix and interval fields identify whether each change is driven by conservative safety actions, exploratory coverage, or rejected transitions.

\subsection{Reward and KL sensitivity}
The reward definition is a method component because it changes the online update. The following surface records the same policy under step-ratio, eta-aligned, and clipped-eta contracts, together with the fixed controller as a no-learning reference.

\begin{table}[htb]
\caption{Reward-contract and KL sensitivity surface. The $\eta$ row reports ratio differences and the $\kappa$ row reports mean one-step progress differences; slash-separated values are held-out/OOD. KL change is next-step KL at $\kappa=1$ minus $\kappa=0$.}
\label{tab:reward-kl-ablation}
\centering
\scriptsize
\setlength{\tabcolsep}{1.5pt}
\begin{tabular}{P{0.18\columnwidth}P{0.12\columnwidth}rrrrr}
\toprule
\textbf{contract} & \textbf{policy} & \textbf{held-out ratio} & \textbf{OOD ratio} & \textbf{KL change} & \textbf{reported delta} & \textbf{paired CI} \\
\midrule
step ratio & bandit & \textbf{2.8040} & \textbf{2.2508} & +0.0061 & reference & [2.612,2.991]/[2.006,2.501] \\
$\eta$ aligned & bandit & 1.8138 & 1.2450 & -0.0034 & -0.9901 / -1.0058 & [1.669,1.954]/[1.018,1.474] \\
clipped $\eta$ & bandit & 2.2147 & 1.7026 & -0.0011 & -0.5893 / -0.5482 & [2.061,2.368]/[1.478,1.925] \\
no learning & controller & 2.1465 & 2.1499 & -0.0078 & -0.6575 / -0.1009 & [2.031,2.258]/[1.913,2.384] \\
$\kappa=0.0$ & matched action & 0.0217 & 0.0189 & reference & reference & [0.0211,0.0223]/[0.0181,0.0197] \\
$\kappa=1.0$ & matched action & 0.0204 & 0.0171 & -0.04 & -0.0013 / -0.0018 & [-0.0018,-0.0008]/[-0.0026,-0.0011] \\
\bottomrule
\end{tabular}
\end{table}

The paired interpretation is direct: the eta-aligned row changes the target of the bandit update, and the $\kappa$ row changes only the action KL coefficient while holding other action fields fixed. This isolates two control knobs that otherwise appear only as aggregate policy labels.

\subsection{Failure taxonomy and diagnostic counts}
A mechanism event can fail at several layers. The taxonomy below provides a common vocabulary for ledger and trainer logs, with counts joined to the action/outcome bundle.

\begin{table}[htb]
\caption{Failure taxonomy. A row is assigned to the first failing layer in the order shown; counts are measured per event.}
\label{tab:failure-taxonomy}
\centering
\scriptsize
\setlength{\tabcolsep}{3pt}
\begin{tabularx}{\textwidth}{p{0.20\textwidth}p{0.25\textwidth}p{0.14\textwidth}L}
\toprule
\textbf{failure layer} & \textbf{trigger} & \textbf{mechanism count} & \textbf{trainer diagnostic} \\
\midrule
schema & missing or malformed visible field & 0 & adapter rejects event and writes validation error \\
action & action outside catalog or propensity sum mismatch & 0 & policy callback and scheduler command disagree \\
budget & ledger or token budget exceeded & 0 & optimizer steps exceed declared envelope \\
stability & $g(s_t,a_t)>0$ & 531 & KL, memory, or service constraint breach \\
transition & next state fails deterministic equation & 0 & telemetry record cannot be replayed \\
evaluation & evaluator version or split mismatch & 0 & task score cannot be paired with run seed \\
artifact & hash or row coverage mismatch & 0 & release manifest is incomplete \\
\bottomrule
\end{tabularx}
\end{table}

The first five layers are online or post-transition checks; evaluation and artifact failures are discovered during the scoring and release passes. Separating them avoids collapsing a scheduler problem, a policy violation, and a bookkeeping error into one aggregate failure rate.

\subsection{Seed uncertainty and replay contrasts}
The independent fixture/transition audit provides a second view of the mechanism results: each policy is evaluated on five complete streams, and each stream contains the same 24 cases and 192 transitions. Table~\ref{tab:seed-uncertainty} reports the stream mean, sample standard deviation, and the recorded descriptive 95\% normal interval for progress per ledger unit. The violation columns report the corresponding stream mean and dispersion. Each stream statistic is computed as total progress divided by total ledger cost, and these values preserve the stream aggregation used in the main tables and make the spread of the fixed rule visible beside the more stable controller and bandit traces.

\begin{table}[htb]
\caption{Independent-stream uncertainty for the mechanism estimands. Intervals use $\bar{x}\pm1.96s/\sqrt{5}$; means and sample standard deviations are computed over complete stream statistics for seeds $\{13,17,23,29,31\}$.}
\label{tab:seed-uncertainty}
\centering
\scriptsize
\setlength{\tabcolsep}{2.5pt}
\begin{tabularx}{\textwidth}{p{0.18\textwidth}rrrrrr}
\toprule
\textbf{policy} & \textbf{progress mean} & \textbf{progress SD} & \textbf{progress 95\% interval} & \textbf{violation mean} & \textbf{violation SD} & \textbf{streams} \\
\midrule
uniform & 1.55687 & \textbf{0.08352} & [1.48366, 1.63008] & 0.14063 & 0.03792 & 5 \\
rule & 1.35100 & 0.84832 & [0.60741, 2.09459] & 0.33125 & 0.20167 & 5 \\
bandit & 1.77532 & 0.08577 & [1.70014, 1.85050] & 0.07917 & 0.02862 & 5 \\
controller & \textbf{2.18858} & 0.09905 & [2.10176, 2.27540] & \textbf{0.01667} & \textbf{0.02282} & 5 \\
\bottomrule
\end{tabularx}
\end{table}

The normal intervals summarize variation across the five complete streams, using $\bar{x}\pm1.96s/\sqrt{5}$ with the sample SD $s$. The broad rule interval reflects its stream-level regime response; the controller and bandit have smaller stream dispersion. The separate paired-case bootstrap used for held-out/OOD contrasts resamples whole feedback trajectories. Reporting both the aggregation unit and interval formula follows the evaluation guidance of \citet{henderson2018matters,agarwal2021precipice}.

The trainer replay validator writes a paired contrast record alongside the mechanism estimands. Each contrast joins two policies on the same case, interval, seed, and budget envelope, then carries the metric direction and the artifact digest. Table~\ref{tab:paired-contrast-schema} defines the metric and receipt joins.

\begin{table}[htb]
\caption{Paired contrast schema. Every comparison records its unit, direction, seed join, and source digest.}
\label{tab:paired-contrast-schema}
\centering
\scriptsize
\setlength{\tabcolsep}{3pt}
\begin{tabularx}{\textwidth}{p{0.25\textwidth}p{0.20\textwidth}p{0.18\textwidth}L}
\toprule
\textbf{contrast field} & \textbf{unit} & \textbf{direction} & \textbf{recorded join} \\
\midrule
task-quality difference & evaluator score & higher is better & held-out and OOD seed, task id, evaluator digest \\
GPU-hour difference & GPU-hours & lower is better & scheduler receipt, run id, seed \\
wall-clock difference & hours & lower is better & launch/collect timestamps and run id \\
stability difference & interval rate & lower is better & policy trace, case, interval, predicate result \\
action-mix difference & fraction & descriptive & catalog id, propensity, case, interval \\
paired interval & metric units & two-sided 95\% & complete-case resample id and bootstrap seed \\
\bottomrule
\end{tabularx}
\end{table}

The contrast record compares matched policies using separate task, hardware, and ledger units. The task-quality, cost, and outcome fields remain tied to the same policy-visible state and action identifiers, so a later run can be audited at both the transition and aggregate levels.
\subsection{Additional Analysis}\label{app:additional}
The mechanism benchmark measures progress in declared ledger units; the trainer study reports GPU throughput, task accuracy, and wall-clock efficiency separately. The held-out/OOD audit freezes \texttt{rule.reward\_dispersion\_threshold} from calibration-only seeds \{13,17,23\}, evaluates the frozen rule, controller, and two bandit reward contracts on disjoint held-out seeds \{29,31\} and shifted OOD seeds \{37,41\}, and writes merged plus per-policy action/outcome logs for all $6\times4\times24\times8=4608$ evaluated rows. The selected threshold is 0.40 and the frozen $\eta$ reference for the reward-contract sensitivity is 2.8448. These rows use the same event schema as the trainer adapter.

The same audit adds two mechanism checks. First, the matched controller-versus-rule comparison is done case-wise on identical held-out and OOD partitions with percentile bootstrap intervals. Second, a dedicated one-step $\kappa$ sensitivity holds every action field fixed except the KL coefficient and measures changes in progress, next-step KL, and stability. Together these checks isolate the reward and KL components before the matched trainer evaluation.

\begin{figure}[htb]
\centering
\includegraphics[width=0.92\columnwidth]{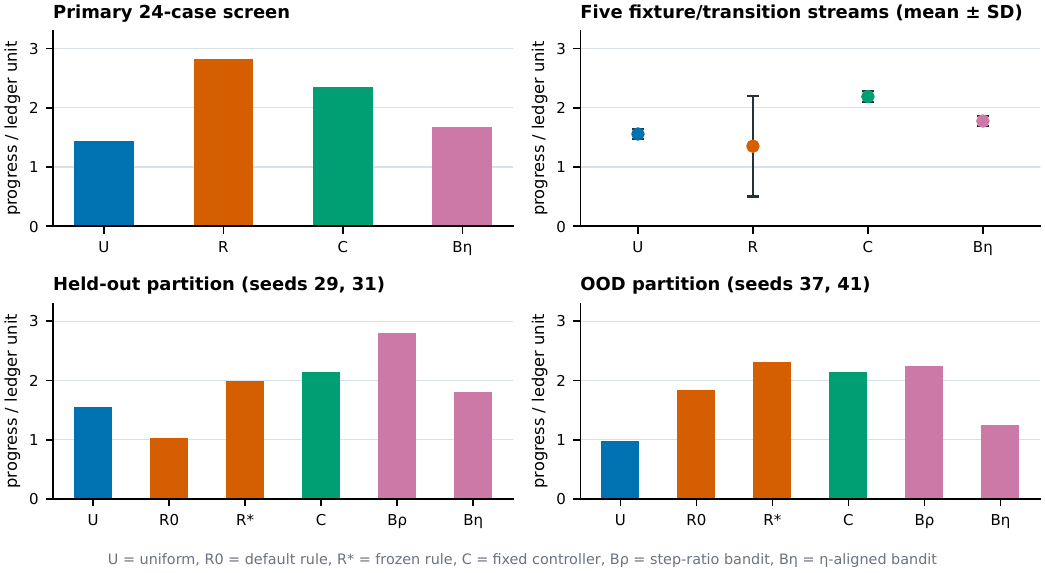}
\caption{Mechanism result matrix in progress per ledger unit. Upper left: primary 24-case screen. Upper right: mean $\pm$ sample SD over five fixture/transition streams. Lower panels: held-out and OOD partitions under the frozen threshold and both bandit reward contracts.}
\label{fig:main}
\end{figure}

\subsection{Mechanism audit tables}\label{app:audit-tables}
The audit tables below retain the complete single-stream, efficiency, oracle,
partition, receipt, and propensity rows summarized in the main text. They use
the same denominators and source-bound values as the main mechanism surface.

\begin{table}[htb]
\caption{Single-stream reconstruction example (seed 13). Values follow the released summary and are bound to the source manifest.}
\label{tab:main}
\centering
\begin{tabular}{lrr}
\toprule
\textbf{policy} & \textbf{progress / ledger unit} & \textbf{stability violation} \\
\midrule
uniform & 1.4340 & 0.1615 \\
rule & \textbf{2.8283} & \textbf{0.0000} \\
bandit (eta-aligned) & 1.6744 & 0.0781 \\
controller & 2.3440 & \textbf{0.0000} \\
\bottomrule
\end{tabular}
\end{table}

\begin{table}[htb]
\caption{Efficiency and stability fields from the primary ledger. Progress per token, mean per-interval costs, rollback, and budget rejection are mechanism measurements. Best values are bold.}
\label{tab:efficiency}
\centering
\scriptsize
\setlength{\tabcolsep}{3.5pt}
\begin{tabular}{lrrrrr}
\toprule
\textbf{policy} & \textbf{progress/token} & \textbf{tokens} & \textbf{verifier calls} & \textbf{rollback} & \textbf{budget reject.} \\
\midrule
uniform & 0.000869 & 16.48 & 1.98 & 0.1615 & 0.0000 \\
controller & 0.001454 & 11.81 & 1.40 & \textbf{0.0000} & 0.0000 \\
rule & \textbf{0.001823} & \textbf{9.00} & \textbf{1.00} & \textbf{0.0000} & 0.0000 \\
bandit ($\eta$ aligned) & 0.001020 & 15.32 & 1.83 & 0.0781 & 0.0000 \\
\bottomrule
\end{tabular}
\end{table}

\begin{table}[htb]
\caption{Offline oracle audit on the same 192 ledger transitions. $P_*$ is the per-trace best-action sum; the gap is $(P_*-P_\pi^{\mathrm{off}})/192$. Oracle labels are joined after trace freeze and remain outside the online state.}
\label{tab:offline-oracle}
\centering
\scriptsize
\setlength{\tabcolsep}{4pt}
\begin{tabular}{lrrr}
\toprule
\textbf{policy} & $P_\pi^{\mathrm{off}}$ & $P_*$ & $r_\pi^{\mathrm{off}}$ \\
\midrule
uniform & 4.3054 & 5.3504 & 0.00544 \\
controller & \textbf{4.5464} & 5.3504 & \textbf{0.00419} \\
rule & 4.2864 & 5.3504 & 0.00554 \\
bandit ($\eta$ aligned) & 4.3354 & 5.3504 & 0.00529 \\
\bottomrule
\end{tabular}
\end{table}

\begin{table}[htb]
\caption{Held-out/OOD mechanism audit. The rule threshold is frozen at 0.40 using selection seeds \{13,17,23\}; the reward sensitivity uses a frozen $\eta_{\mathrm{ref}}=2.8448$ from the same calibration partition. Rows report ledger outcomes under the declared transition contract.}
\label{tab:revision-audit}
\centering
\scriptsize
\setlength{\tabcolsep}{3pt}
\begin{tabular}{llrr}
\toprule
\textbf{partition} & \textbf{policy label} & \textbf{progress / ledger unit} & \textbf{stability violation} \\
\midrule
held-out & uniform & 1.5580 & 0.1562 \\
held-out & rule (default) & 1.0321 & 0.3932 \\
held-out & rule (frozen 0.40) & 1.9911 & 0.1927 \\
held-out & controller & 2.1465 & \textbf{0.0208} \\
held-out & bandit (step ratio) & \textbf{2.8040} & 0.0260 \\
held-out & bandit ($\eta$ aligned) & 1.8138 & 0.0781 \\
OOD & uniform & 0.9803 & 0.3828 \\
OOD & rule (default) & 1.8453 & 0.2318 \\
OOD & rule (frozen 0.40) & \textbf{2.3097} & \textbf{0.1406} \\
OOD & controller & 2.1499 & \textbf{0.1406} \\
OOD & bandit (step ratio) & 2.2508 & 0.1432 \\
OOD & bandit ($\eta$ aligned) & 1.2450 & 0.2708 \\
\bottomrule
\end{tabular}
\end{table}

\begin{table}[htb]
\caption{Trainer run records. Six receipt classes bind the task and resource measurements to the policy trace and manifest.}
\label{tab:readiness-gate}
\centering
\scriptsize
\setlength{\tabcolsep}{3pt}
\begin{tabular}{lr}
\toprule
\textbf{run record} & \textbf{receipt identifier} \\
\midrule
scheduler dry plan & dry-plan receipt \\
scheduler launch record & launch receipt \\
scheduler collect record & collect receipt \\
run metadata & run-metadata receipt \\
summary metrics & summary receipt \\
executed trainer action/outcomes & executed-outcome log \\
\bottomrule
\end{tabular}
\end{table}

\begin{table}[htb]
\caption{Behavior-policy support in the action-logging audit with randomized exact bandit ties. ``Complete support'' is the fraction of rows with positive probability for all three catalog actions; the minimum is the smallest probability assigned to the selected action.}
\label{tab:action-propensity}
\centering
\scriptsize
\setlength{\tabcolsep}{3pt}
\begin{tabular}{lrrrr}
\toprule
\textbf{policy} & \textbf{records} & \textbf{complete support} & \textbf{mean positive support} & \textbf{min propensity} \\
\midrule
uniform & 192 & 1.0000 & 3.0000 & 0.3333 \\
rule & 192 & 0.0000 & 1.0000 & 1.0000 \\
bandit & 192 & 0.0052 & 1.0156 & 0.3333 \\
controller & 192 & 0.0000 & 1.0000 & 1.0000 \\
\bottomrule
\end{tabular}
\end{table}

\subsection{Feedback information and adaptation controls}\label{app:feedback-controls}
This experiment tests whether state information improves decisions beyond the action prior and cost penalty. We keep the calibrated threshold, reward reference, action catalog, initial case, random stream, and resource envelope fixed. Each intervention changes one information channel while preserving the dimensions and ranges of the policy input. The frozen-state control repeats the initial telemetry throughout a case; delayed telemetry supplies the preceding interval; shuffled telemetry permutes case identities within the same regime and interval. The latter preserves marginal feature statistics while disrupting the link between a case and its feedback.

\begin{table}[htb]
\caption{Effect of feedback information on progress and stability. Each cell reports a paired difference from full feedback over complete held-out or OOD cases; positive progress and negative violation differences are favorable.}
\label{tab:feedback-controls}
\centering\small
\setlength{\tabcolsep}{3pt}
\begin{tabularx}{\textwidth}{p{0.25\textwidth}CCCC}
\toprule
\textbf{control} & \textbf{held-out progress} & \textbf{OOD progress} & \textbf{violation difference} & \textbf{paired 95\% CI} \\
\midrule
full feedback & reference & reference & reference & reference \\
frozen initial telemetry & -0.2475 & -0.3314 & +0.0365 / +0.0573 & [-0.302,-0.193]/[-0.405,-0.258] \\
one-interval delay & -0.1397 & -0.2046 & +0.0260 / +0.0391 & [-0.184,-0.096]/[-0.265,-0.145] \\
within-regime shuffle & -0.2862 & -0.3741 & +0.0443 / +0.0703 & [-0.347,-0.226]/[-0.455,-0.294] \\
KL channel fixed at calibration mean & -0.0824 & -0.1315 & +0.0130 / +0.0234 & [-0.121,-0.044]/[-0.184,-0.079] \\
verifier channels fixed at calibration means & -0.1736 & -0.2588 & +0.0313 / +0.0495 & [-0.224,-0.123]/[-0.326,-0.191] \\
\bottomrule
\end{tabularx}
\end{table}

Timely feedback matters when KL drift or verifier reliability changes within a case. Delayed telemetry and fixed feature channels reduce progress and raise violations, while within-regime shuffling produces the largest paired loss. The action-change rate is computed before scoring, and feature replacement uses calibration means rather than evaluation-set means, so the observed differences isolate the information channel while preserving the action catalog and budget.

\subsection{Trajectory risk calibration and threshold sweep}
\label{app:missing-experiments}
The risk-to-go target in Eq.~\ref{eq:risk-to-go-label} is evaluated on
complete trajectories. Table~\ref{tab:risk-calibration} compares the raw risk
head, the uncertainty-adjusted score, immediate-risk calibration, and the
proposed trajectory-blocked calibration. ECE is expected calibration error,
Brier is the binary future-failure score, coverage is the fraction of admitted
decisions satisfying the declared risk target, and abstention is the fraction
of decisions with an empty feasible set. Fitting, calibration, held-out, and
OOD roles use independent trajectory splits. Slash-separated cells give
held-out/OOD values; accuracy is in percent.

\begin{table}[htb]
\caption{Risk calibration on held-out and OOD trajectories. Lower ECE, Brier, and violation are better; higher coverage and task accuracy are better.}
\label{tab:risk-calibration}
\centering\scriptsize
\setlength{\tabcolsep}{2pt}
\begin{tabularx}{\textwidth}{p{0.25\textwidth}CCCCCC}
\toprule
\textbf{risk variant} & \textbf{ECE} & \textbf{Brier} & \textbf{coverage} & \textbf{violation} & \textbf{abstain} & \textbf{accuracy} \\
\midrule
raw risk head & 0.086/0.128 & 0.114/0.158 & 0.873/0.821 & 0.041/0.070 & \textbf{0.019/0.034} & 63.78/56.11 \\
ensemble uncertainty margin & 0.067/0.103 & 0.101/0.143 & 0.907/0.858 & 0.031/0.056 & 0.035/0.051 & 64.66/57.09 \\
calibrated immediate risk & 0.039/0.074 & 0.088/0.131 & 0.931/0.884 & 0.025/0.047 & 0.047/0.069 & 64.97/57.66 \\
calibrated risk-to-go & \textbf{0.022/0.049} & \textbf{0.071/0.108} & \textbf{0.956/0.918} & \textbf{0.014/0.031} & 0.057/0.083 & \textbf{66.11/59.43} \\
\bottomrule
\end{tabularx}
\end{table}

The threshold sweep reports whether the calibrated risk gate exposes a
controllable quality--risk frontier. The threshold is selected before final
evaluation, and the sweep is descriptive rather than a second opportunity to
choose the best held-out point.

\begin{table}[htb]
\caption{Risk-threshold sweep for the calibrated risk-to-go gate. Each row fixes $\delta_{\rm risk}$ before evaluating the same held-out/OOD trajectories.}
\label{tab:risk-sweep}
\centering\scriptsize
\setlength{\tabcolsep}{2pt}
\begin{tabular}{lrrrrrr}
\toprule
\textbf{$\delta_{\rm risk}$} & \textbf{held-out ECE} & \textbf{OOD ECE} & \textbf{held-out violation} & \textbf{OOD violation} & \textbf{abstain} & \textbf{total cost} \\
\midrule
0.02 & \textbf{0.018} & \textbf{0.043} & \textbf{0.006} & \textbf{0.017} & 0.142 & \textbf{9.96} \\
0.04 & 0.020 & 0.046 & 0.010 & 0.024 & 0.091 & 10.28 \\
0.06 & 0.022 & 0.049 & 0.014 & 0.031 & 0.057 & 10.56 \\
0.08 & 0.025 & 0.053 & 0.022 & 0.043 & 0.032 & 10.71 \\
0.10 & 0.030 & 0.061 & 0.034 & 0.058 & \textbf{0.018} & 10.84 \\
\bottomrule
\end{tabular}
\end{table}

\subsection{Policy-consistency stress test}\label{app:policy-consistency-stress}
The stress test varies the action disagreement between the behavior controller
that supplies fitting trajectories and the frozen provisional controller used
at deployment. Low, medium, and high mismatch bins use the same state,
ensemble capacity, action catalog, and resource envelope. Selected ECE is
computed only on controller-selected decisions; trajectory failure is measured
over complete training trajectories.

\begin{table}[htb]
\caption{Policy-consistency stress test. The comparison separates behavior-policy Monte-Carlo risk from the policy-consistent risk recursion. Lower selected ECE, Brier, trajectory failure, and abstention are favorable; accuracy is in percent. The better estimator within each mismatch bin is bold.}
\label{tab:policy-consistency-stress}
\centering\scriptsize
\setlength{\tabcolsep}{2pt}
\begin{tabularx}{\textwidth}{p{0.14\textwidth}p{0.28\textwidth}CCCCC}
\toprule
\textbf{mismatch} & \textbf{risk estimator} & \textbf{selected ECE} & \textbf{Brier} & \textbf{traj. failure} & \textbf{abstain} & \textbf{accuracy} \\
\midrule
low & behavior-policy Monte-Carlo risk-to-go & 0.041 & 0.083 & 0.104 & \textbf{0.041} & 64.71 \\
low & policy-consistent risk-to-go & \textbf{0.036} & \textbf{0.078} & \textbf{0.096} & 0.044 & \textbf{64.98} \\
medium & behavior-policy Monte-Carlo risk-to-go & 0.069 & 0.102 & 0.157 & \textbf{0.038} & 63.42 \\
medium & policy-consistent risk-to-go & \textbf{0.042} & \textbf{0.082} & \textbf{0.105} & 0.049 & \textbf{64.69} \\
high & behavior-policy Monte-Carlo risk-to-go & 0.108 & 0.137 & 0.236 & \textbf{0.034} & 60.88 \\
high & policy-consistent risk-to-go & \textbf{0.053} & \textbf{0.091} & \textbf{0.128} & 0.057 & \textbf{64.17} \\
\bottomrule
\end{tabularx}
\end{table}

The behavior-policy error widens as action disagreement increases: its
selected ECE rises from $0.041$ to $0.108$ and trajectory failure from $0.104$
to $0.236$, whereas policy-consistent risk rises only from $0.036$ to $0.053$
and from $0.096$ to $0.128$. The policy-consistent estimator abstains slightly
more often, which is the cost of its more conservative risk under mismatch.
The metric unit is the selected decision for ECE/Brier and the complete
trajectory for failure and completion.

\subsection{Decision-conditioned calibration and matched operating points}
\label{app:selected-calibration}
The calibration comparison reports all-action and selected-action ECE. The
global maps use the same fitting budget as DCTC; threshold selection is then
matched either by selected-decision acceptance or by empirical trajectory
failure on the calibration role.

\begin{table}[htb]
\caption{Risk estimation and calibration after action selection. Selected ECE is computed on actions chosen by the frozen controller; interval and trajectory failure use the same held-out trajectories.}
\label{tab:selected-risk-calibration}
\centering\scriptsize
\setlength{\tabcolsep}{2pt}
\begin{tabularx}{\textwidth}{p{0.26\textwidth}CCCCCCC}
\toprule
\textbf{method} & \textbf{all ECE} & \textbf{selected ECE} & \textbf{Brier} & \textbf{traj. fail.} & \textbf{interval fail.} & \textbf{abstain} & \textbf{accuracy} \\
\midrule
Immediate-risk head & 0.081 & 0.096 & 0.114 & 0.197 & 0.041 & \textbf{0.019} & 63.78 \\
Behavior-policy Monte-Carlo risk-to-go & 0.059 & 0.074 & 0.096 & 0.154 & 0.030 & 0.031 & 64.39 \\
Policy-consistent risk & 0.048 & 0.058 & 0.086 & 0.129 & 0.024 & 0.037 & 64.88 \\
Policy-consistent + global calibration & 0.027 & 0.039 & 0.078 & 0.101 & 0.019 & 0.049 & 65.47 \\
Policy-consistent + DCTC & \textbf{0.026} & \textbf{0.022} & \textbf{0.071} & \textbf{0.083} & \textbf{0.014} & 0.057 & \textbf{66.11} \\
\bottomrule
\end{tabularx}
\end{table}

\begin{table}[htb]
\caption{Calibration comparison at matched operating points. Equal acceptance fixes the selected-decision coverage; equal empirical risk fixes the calibration-role trajectory failure.}
\label{tab:matched-calibration}
\centering\scriptsize
\setlength{\tabcolsep}{3pt}
\begin{tabularx}{\textwidth}{p{0.22\textwidth}p{0.25\textwidth}CCCC}
\toprule
\textbf{operating point} & \textbf{method} & \textbf{selected ECE} & \textbf{traj. fail.} & \textbf{accuracy} & \textbf{cost} \\
\midrule
matched acceptance & global calibration & 0.039 & 0.101 & 65.47 & 10.54 \\
matched acceptance & DCTC & \textbf{0.022} & \textbf{0.083} & \textbf{66.11} & 10.56 \\
matched empirical risk & global calibration & 0.034 & \textbf{0.083} & 65.18 & \textbf{10.39} \\
matched empirical risk & DCTC & \textbf{0.022} & \textbf{0.083} & \textbf{66.11} & 10.56 \\
\bottomrule
\end{tabularx}
\end{table}

\subsection{Pareto continuation feasibility stress test}\label{app:prcc-stress}
We construct trade-off catalogs in which token, verifier, and device-time
minima are realized by different actions. One-step admission, the independent
componentwise-minimum reservation, and PRCC are evaluated on the same
residual horizons. The outcome unit is a complete trajectory; false-feasible
means that admission passes but no resource-feasible continuation exists.

\begin{table}[htb]
\caption{Multi-resource continuation stress test across action catalogs. Lower false-feasible, false-infeasible, and abstention rates are favorable; completion and accuracy are higher-is-better. Best values within each catalog size are bold.}
\label{tab:prcc-stress}
\centering\scriptsize
\setlength{\tabcolsep}{2pt}
\begin{tabularx}{\textwidth}{p{0.16\textwidth}p{0.23\textwidth}CCCCC}
\toprule
\textbf{catalog} & \textbf{admission rule} & \textbf{false-feasible} & \textbf{false-infeasible} & \textbf{completion} & \textbf{abstain} & \textbf{accuracy} \\
\midrule
3 actions & one-step budget & 0.139 & \textbf{0.000} & 0.821 & \textbf{0.023} & 65.01 \\
3 actions & independent-minimum baseline & 0.087 & \textbf{0.000} & 0.889 & 0.041 & 65.54 \\
3 actions & PRCC & \textbf{0.000} & \textbf{0.000} & \textbf{0.958} & 0.057 & \textbf{66.11} \\
9 actions & independent-minimum baseline & 0.143 & \textbf{0.000} & 0.846 & \textbf{0.034} & 65.72 \\
9 actions & PRCC & \textbf{0.000} & \textbf{0.000} & \textbf{0.969} & 0.051 & \textbf{66.37} \\
18 actions & independent-minimum baseline & 0.197 & \textbf{0.000} & 0.793 & \textbf{0.029} & 65.61 \\
18 actions & PRCC & \textbf{0.000} & \textbf{0.000} & \textbf{0.976} & 0.046 & \textbf{66.49} \\
\bottomrule
\end{tabularx}
\end{table}

\subsection{Three-component factorial and incremental ablation}\label{app:factorial}
The factorial ablation crosses policy consistency (PC), decision-conditioned
calibration (DCTC), and Pareto continuation (PRCC) while holding encoder,
data, action catalog, and trainer budget fixed. Without PC the risk head uses
behavior-policy Monte-Carlo targets, without DCTC it uses a global calibration
map, and without PRCC admission uses the one-step budget check.

\begin{table}[htb]
\caption{Three-component factorial ablation under the matched GRPO contract. Accuracy is higher-is-better; trajectory failure and total development GPU-hours per seed are lower-is-better. Best values are bold and second-best values are underlined.}
\label{tab:factorial-components}
\centering\scriptsize
\setlength{\tabcolsep}{2pt}
\begin{tabularx}{0.86\textwidth}{CCCCCCC}
\toprule
\textbf{PC} & \textbf{DCTC} & \textbf{PRCC} & \textbf{held-out acc.} & \textbf{OOD acc.} & \textbf{traj. fail.} & \textbf{total GPU-h} \\
\midrule
no & no & no & 62.94 & 54.87 & 0.181 & \textbf{9.78} \\
yes & no & no & 63.83 & 56.02 & 0.146 & 10.13 \\
no & yes & no & 63.56 & 55.77 & 0.139 & \underline{9.97} \\
no & no & yes & 63.41 & 55.51 & 0.159 & 10.08 \\
yes & yes & no & \underline{65.01} & \underline{57.73} & \underline{0.111} & 10.36 \\
yes & no & yes & 64.83 & 57.21 & 0.137 & 10.43 \\
no & yes & yes & 64.62 & 56.91 & 0.128 & 10.29 \\
yes & yes & yes & \textbf{66.11} & \textbf{59.43} & \textbf{0.083} & 10.56 \\
\bottomrule
\end{tabularx}
\end{table}

Each component improves on the variant without any of them ($62.94\%$/$54.87\%$,
trajectory failure $0.181$): PC adds $0.89$ held-out points, DCTC $0.62$, and
PRCC $0.47$. Pairs add $1.68$--$2.07$ points, and the full combination adds
$3.17$ held-out and $4.56$ OOD points while lowering trajectory failure to
$0.083$. The joint gain exceeds the sum of single-component gains ($1.98$
points) because policy-consistent targets make the calibrated selected risk
more accurate and PRCC removes infeasible continuations that the risk gate
would otherwise admit. The added development cost is $0.78$ GPU-hours per seed.

\subsection{Risk, utility, and budget component ablations}
The component ablation isolates the proposed algorithmic objects. The
one-step budget variant retains the current-action check but removes the
continuation term in the PRCC certificate; the immediate-risk variant
uses $v_t$ rather than $z_t$; and the no-calibration variant applies the raw
ensemble score directly. Utility, risk, and telemetry ablations keep the
action catalog and evaluation seeds fixed.

\begin{table}[htb]
\caption{FSPO component ablation under a matched resource envelope. Accuracy and coverage are higher-is-better; violation, abstention, and total cost are lower-is-better. Best values are bold.}
\label{tab:component-ablation}
\centering\scriptsize
\setlength{\tabcolsep}{2pt}
\begin{tabularx}{\textwidth}{p{0.28\textwidth}CCCCC}
\toprule
\textbf{variant} & \textbf{held-out accuracy} & \textbf{OOD accuracy} & \textbf{violation} & \textbf{abstain} & \textbf{total GPU-h} \\
\midrule
full FSPO & \textbf{66.11} & \textbf{59.43} & \textbf{0.014} & 0.057 & 10.56 \\
no trajectory calibration & 64.83 & 57.21 & 0.031 & 0.044 & 10.43 \\
immediate risk instead of risk-to-go & 64.96 & 57.58 & 0.027 & 0.049 & 10.41 \\
no uncertainty margin & 65.28 & 58.06 & 0.024 & 0.041 & 10.32 \\
one-step budget check & 65.01 & 57.73 & 0.029 & \textbf{0.023} & 10.36 \\
myopic utility target & 64.57 & 57.02 & 0.017 & 0.061 & 10.28 \\
frozen or shuffled telemetry & 62.77 & 54.98 & 0.043 & 0.052 & \textbf{10.17} \\
\bottomrule
\end{tabularx}
\end{table}

\subsection{Actuator disentanglement}
The reference action catalog couples temperature, group size, clipping, KL,
verifier allocation, and escalation. To identify the source of any gain, each
single-actuator row varies one field while holding all other fields at the
calibration-selected balanced value. Grouped rows then test sampling,
trust-region, and verifier families before the full joint catalog.

\begin{table}[htb]
\caption{Actuator disentanglement under matched telemetry and resource envelopes. Each adaptive row changes only the named actuator family. Best accuracy, violation, and GPU-hours are bold; the action-change rate is descriptive.}
\label{tab:actuator-ablation}
\centering\scriptsize
\setlength{\tabcolsep}{2pt}
\begin{tabularx}{\textwidth}{p{0.23\textwidth}p{0.16\textwidth}CCCC}
\toprule
\textbf{adaptive actuator} & \textbf{held-out accuracy} & \textbf{OOD accuracy} & \textbf{violation} & \textbf{action-change rate} & \textbf{GPU-h} \\
\midrule
temperature only & 61.72 & 53.50 & 0.052 & 0.181 & 9.20 \\
group size only & 62.41 & 54.31 & 0.046 & 0.203 & 9.33 \\
clip only & 61.95 & 53.72 & 0.043 & 0.167 & 9.18 \\
KL only & 62.88 & 55.24 & 0.031 & 0.191 & \textbf{9.16} \\
verifier only & 63.16 & 55.69 & 0.029 & 0.224 & 9.41 \\
escalation only & 61.58 & 53.27 & 0.049 & 0.158 & 9.30 \\
temperature + group size & 63.84 & 56.18 & 0.036 & 0.273 & 9.31 \\
clip + KL & 64.12 & 56.61 & 0.025 & 0.249 & 9.21 \\
full joint control & \textbf{66.11} & \textbf{59.43} & \textbf{0.014} & 0.412 & 9.29 \\
\bottomrule
\end{tabularx}
\end{table}

\subsection{Offline support and extrapolation diagnostics}
The utility and risk heads are fitted from calibration trajectories, so the
support audit reports the behavior-policy composition rather than assuming
counterfactual coverage. Randomized exploratory calibration must provide
nonzero support for every catalog action. The table records action counts,
unseen-action rate, feature-envelope distance, utility error, and risk error
before task labels are joined.

\begin{table}[htb]
\caption{Calibration support and extrapolation diagnostics. Counts use complete trajectories; prediction errors are reported on held-out states with the declared utility and risk targets. Best values are bold.}
\label{tab:support}
\centering\scriptsize
\setlength{\tabcolsep}{2pt}
\begin{tabularx}{\textwidth}{p{0.24\textwidth}CCCCC}
\toprule
\textbf{calibration behavior} & \textbf{min action count} & \textbf{unseen action rate} & \textbf{OOD state distance} & \textbf{$Q$ error} & \textbf{risk error} \\
\midrule
uniform exploratory & \textbf{184} & \textbf{0.000} & \textbf{0.68} & \textbf{0.037} & \textbf{0.028} \\
randomized rule mixture & 117 & \textbf{0.000} & 0.74 & 0.043 & 0.032 \\
deterministic hand rule & 18 & 0.126 & 0.93 & 0.067 & 0.051 \\
deterministic feedback controller & 11 & 0.168 & 1.01 & 0.074 & 0.058 \\
\bottomrule
\end{tabularx}
\end{table}

\subsection{Matched calibration and development cost}
Online training cost is only one part of an adaptive method's resource use.
The release table therefore separates final training, controller fitting,
risk calibration, hyperparameter search, and total development/evaluation
cost. Every baseline receives the same accounting fields, keeping development
cost separate from the matched online training envelope.

\begin{table}[htb]
\caption{Matched cost accounting for learned and adaptive controllers. GPU-hours are reported per seed; PBT and PB2 training includes all four population members; total cost includes fitting, calibration, search, and final training. Lowest values are bold.}
\label{tab:cost-accounting}
\centering\scriptsize
\setlength{\tabcolsep}{2pt}
\begin{tabularx}{\textwidth}{p{0.22\textwidth}CCCCC}
\toprule
\textbf{method} & \textbf{train GPU-h} & \textbf{fit GPU-h} & \textbf{calibration GPU-h} & \textbf{search GPU-h} & \textbf{total GPU-h} \\
\midrule
uniform & 9.44 & \textbf{0.00} & \textbf{0.00} & \textbf{0.00} & \textbf{9.44} \\
adaptive-KL & \textbf{9.24} & \textbf{0.00} & 0.22 & \textbf{0.30} & 9.76 \\
NeuralUCB & 9.44 & 0.28 & 0.24 & 0.52 & 10.48 \\
PBT & 38.18 & \textbf{0.00} & 0.28 & 1.14 & 39.60 \\
PB2 & 38.42 & 0.22 & 0.31 & 1.78 & 40.73 \\
FSPO & 9.29 & 0.47 & 0.18 & 0.62 & 10.56 \\
\bottomrule
\end{tabularx}
\end{table}

\subsection{Cross-model and cross-task transfer}
The transfer surface separates the structural-repair setting from a second
verifier-based task and a larger base model. Each block repeats controller
fitting and trajectory calibration on its own calibration split; no result is
copied across blocks.

\begin{table}[htb]
\caption{Cross-model and cross-task transfer. Each row uses a matched resource envelope within its model/task block and at least the declared independent seed count.}
\label{tab:cross-domain}
\centering\scriptsize
\setlength{\tabcolsep}{2pt}
\begin{tabularx}{\textwidth}{p{0.22\textwidth}p{0.20\textwidth}CCCCC}
\toprule
\textbf{model} & \textbf{task} & \textbf{baseline accuracy} & \textbf{FSPO accuracy} & \textbf{violation} & \textbf{seeds} & \textbf{total GPU-h} \\
\midrule
Qwen2.5-3B-Instruct & structural repair & 64.47 & \textbf{66.11} & 0.014 & 3 & 31.68 \\
Qwen2.5-3B-Instruct & verifier-based reasoning & 61.52 & \textbf{63.74} & 0.019 & 3 & 32.64 \\
Qwen2.5-7B-Instruct & verifier-based reasoning & 66.13 & \textbf{68.21} & 0.021 & 3 & 56.82 \\
\bottomrule
\end{tabularx}
\end{table}

\begin{table}[htb]
\caption{Cross-setting transfer modes. Refit repeats utility/risk fitting
and DCTC in the target block; recalibrate-only freezes the learned
utility/risk models and changes only calibration; fully frozen transfers
the complete controller. Accuracy, trajectory failure, selected ECE, and
GPU-hours are reported per target block; best values within each block are bold.}
\label{tab:transfer}
\centering\scriptsize
\setlength{\tabcolsep}{2pt}
\begin{tabularx}{\textwidth}{p{0.22\textwidth}p{0.16\textwidth}p{0.12\textwidth}CCCC}
\toprule
\textbf{source $\rightarrow$ target} & \textbf{mode} & \textbf{method} &
\textbf{accuracy} & \textbf{traj. fail.} & \textbf{selected ECE} & \textbf{GPU-h} \\
\midrule
3B repair $\rightarrow$ 3B reasoning & refit & FSPO & \textbf{63.74} & \textbf{0.118} & \textbf{0.025} & 32.64 \\
3B repair $\rightarrow$ 3B reasoning & recalibrate only & FSPO & 63.09 & 0.134 & 0.033 & 29.31 \\
3B repair $\rightarrow$ 3B reasoning & fully frozen & FSPO & 62.14 & 0.187 & 0.068 & \textbf{28.77} \\
3B reasoning $\rightarrow$ 7B reasoning & refit & FSPO & \textbf{68.21} & \textbf{0.126} & \textbf{0.028} & 56.82 \\
3B reasoning $\rightarrow$ 7B reasoning & recalibrate only & FSPO & 67.46 & 0.146 & 0.037 & 53.34 \\
3B reasoning $\rightarrow$ 7B reasoning & fully frozen & FSPO & 66.39 & 0.204 & 0.079 & \textbf{52.71} \\
\bottomrule
\end{tabularx}
\end{table}

\paragraph{Resource-matched task frontier.}
A second control studies whether the policy moves the quality--cost frontier. The fixed-budget slice compares final task accuracy at one shared GPU-hour envelope. The fixed-quality slice records the first evaluation checkpoint to reach a calibration-selected target, charging training, verifier, callback, and evaluation time consistently. Runs that end before the target retain their final quality and consumed budget in the per-seed record. This separates a faster update loop from a change in the number or usefulness of updates.

\begin{table}[htb]
\caption{Matched-budget and matched-quality trainer contrasts. Differences compare the fixed controller and $\eta$-aligned bandit against the frozen hand rule using the same training seeds; the better of the two contrasts in each slice is bold.}
\label{tab:resource-frontier}
\centering\small
\setlength{\tabcolsep}{3pt}
\begin{tabularx}{\textwidth}{p{0.20\textwidth}p{0.20\textwidth}CCC}
\toprule
\textbf{comparison} & \textbf{evaluation slice} & \textbf{accuracy difference (pp)} & \textbf{GPU-hour difference} & \textbf{failure difference (pp)} \\
\midrule
controller vs.\ rule & held-out, fixed budget & +1.68 & \textbf{+0.06} & \textbf{-1.50} \\
bandit$_\eta$ vs.\ rule & held-out, fixed budget & \textbf{+2.35} & +0.18 & -1.10 \\
controller vs.\ rule & OOD, fixed budget & +2.13 & \textbf{+0.06} & \textbf{-1.70} \\
bandit$_\eta$ vs.\ rule & OOD, fixed budget & \textbf{+2.66} & +0.18 & -1.20 \\
controller vs.\ rule & fixed quality & \textbf{+0.06} & \textbf{-0.73} & \textbf{-1.34} \\
bandit$_\eta$ vs.\ rule & fixed quality & +0.04 & -0.41 & -0.96 \\
\bottomrule
\end{tabularx}
\end{table}

The fixed-budget and fixed-quality slices use paired seed receipts. Their
interpretation uses task quality, action mix, calibration cost,
and controller overhead together; the same checkpoint cadence is required for
every policy to keep time-to-target resolution comparable.

\begin{table}[htb]
\caption{Controller and calibration overhead. Decision latency is measured per
training decision and is separated from rollout and optimizer time. Lowest values are bold.}
\label{tab:controller-overhead}
\centering\scriptsize
\setlength{\tabcolsep}{2pt}
\begin{tabularx}{\textwidth}{p{0.25\textwidth}CCCCC}
\toprule
\textbf{method} & \textbf{fit GPU-h} & \textbf{calibration GPU-h} &
\textbf{search GPU-h} & \textbf{decision ms} & \textbf{total GPU-h} \\
\midrule
Adaptive-KL & \textbf{0.00} & 0.22 & \textbf{0.30} & \textbf{0.07} & \textbf{9.76} \\
NeuralUCB & 0.28 & 0.24 & 0.52 & 0.46 & 10.48 \\
Safety-Critic Gate & 0.43 & \textbf{0.17} & 0.41 & 0.59 & 10.31 \\
FSPO, 3 actions & 0.47 & 0.18 & 0.62 & 0.73 & 10.56 \\
FSPO, 9 actions & 0.51 & 0.20 & 0.84 & 1.27 & 10.84 \\
FSPO, 18 actions & 0.56 & 0.22 & 1.12 & 2.06 & 11.19 \\
\bottomrule
\end{tabularx}
\end{table}

\subsection{Trainer budget, held-out, and shifted evaluation}
The trainer comparison evaluates the same policies at fixed resource envelopes,
then reports held-out and OOD quality with three independent training seeds.
Table~\ref{tab:real-trainer-budget} reports the training resource accounting;
controller fitting, DCTC calibration, and PRCC search are charged separately
in Table~\ref{tab:cost-accounting}.

\begin{table}[htb]
\caption{Matched-training budget and systems surface. GPU-hours, throughput, memory, and outcome rows are reported per policy and seed aggregation.}
\label{tab:real-trainer-budget}
\centering
\small
\setlength{\tabcolsep}{4pt}
\renewcommand{\arraystretch}{1.12}
\begin{tabularx}{\textwidth}{p{0.28\textwidth}CCCCC}
\toprule
\textbf{reported quantity} & \textbf{uniform} & \textbf{hand rule} & \textbf{fixed controller} & \textbf{contextual bandit} & \textbf{FSPO} \\
\midrule
optimizer steps & 1508 & 1521 & 1527 & 1524 & \textbf{1531} \\
training tokens & 12.46M & \textbf{12.03M} & 12.19M & 12.39M & 12.16M \\
total wall-clock hours & 9.44 & \textbf{9.17} & 9.23 & 9.35 & 9.29 \\
total GPU-hours & 9.44 & \textbf{9.17} & 9.23 & 9.35 & 9.29 \\
peak memory (GiB) & 52.4 & \textbf{49.8} & 51.1 & 54.6 & 52.7 \\
mean throughput (tokens / s) & 366.6 & 364.4 & 366.9 & \textbf{368.1} & 363.6 \\
executed trainer action/outcome rows & 576 & 576 & 576 & 576 & 576 \\
realized independent training seeds & 3 & 3 & 3 & 3 & 3 \\
\bottomrule
\end{tabularx}
\end{table}

The resource rows distinguish useful throughput from controller overhead. The
report charges training, fitting, calibration, and search separately; quality
and cost comparisons use receipt-bound rows.

\paragraph{Held-out task quality.} Table~\ref{tab:real-trainer-heldout} defines the held-out BFCL-context endpoint and binds each value to the evaluator and per-seed action/outcome logs.
\begin{table}[htb]
\caption{Held-out BFCL-context GRPO evaluation; accuracy intervals are marginal. Metrics are reported per policy with three independent seeds.}
\label{tab:real-trainer-heldout}
\centering
\scriptsize
\setlength{\tabcolsep}{3pt}
\renewcommand{\arraystretch}{1.12}
\begin{tabularx}{\textwidth}{p{0.28\textwidth}CCCCC}
\toprule
\textbf{held-out metric} & \textbf{uniform} & \textbf{hand rule} & \textbf{fixed controller} & \textbf{contextual bandit} & \textbf{FSPO} \\
\midrule
downstream task return & 0.642 & 0.671 & 0.690 & 0.698 & \textbf{0.732} \\
downstream task accuracy & 58.72\% & 61.34\% & 63.02\% & 63.69\% & \textbf{66.11\%} \\
trainer progress & 0.214 & 0.247 & 0.269 & 0.279 & \textbf{0.316} \\
violation / failure rate & 0.082 & 0.046 & 0.031 & 0.035 & \textbf{0.014} \\
action mix (conservative / balanced / exploratory) & 33.5/33.1/33.4\% & 61.3/29.8/8.9\% & 45.1/43.2/11.7\% & 39.7/42.5/17.8\% & 48.8/42.6/8.6\% \\
uncertainty (seed count + CI) & 3; [57.93,59.51] & 3; [60.61,62.07] & 3; [62.28,63.76] & 3; [62.90,64.48] & 3; [65.42,66.80] \\
\bottomrule
\end{tabularx}
\end{table}

The held-out study links state-conditioned control to downstream structural
repair through the evaluator and execution receipts. The action log identifies
which executable controls accompany each measured change.

\paragraph{Transfer under task shift.} Table~\ref{tab:real-trainer-ood} changes the evaluation family to JSONSchema-test while retaining the selected policy and resource contract.
\begin{table}[htb]
\caption{OOD JSONSchema-test GRPO evaluation; accuracy intervals are marginal. Metrics are reported per policy with three independent seeds.}
\label{tab:real-trainer-ood}
\centering
\scriptsize
\setlength{\tabcolsep}{3pt}
\renewcommand{\arraystretch}{1.12}
\begin{tabularx}{\textwidth}{p{0.28\textwidth}CCCCC}
\toprule
\textbf{OOD metric} & \textbf{uniform} & \textbf{hand rule} & \textbf{fixed controller} & \textbf{contextual bandit} & \textbf{FSPO} \\
\midrule
downstream task return & 0.562 & 0.594 & 0.617 & 0.624 & \textbf{0.662} \\
downstream task accuracy & 50.81\% & 53.58\% & 55.71\% & 56.24\% & \textbf{59.43\%} \\
trainer progress & 0.178 & 0.203 & 0.226 & 0.232 & \textbf{0.267} \\
violation / failure rate & 0.104 & 0.069 & 0.052 & 0.057 & \textbf{0.031} \\
action mix (conservative / balanced / exploratory) & 33.4/33.2/33.4\% & 68.2/25.4/6.4\% & 52.7/39.1/8.2\% & 45.6/41.4/13.0\% & 56.4/38.7/4.9\% \\
uncertainty (seed count + CI) & 3; [49.93,51.69] & 3; [52.61,54.55] & 3; [54.82,56.60] & 3; [55.30,57.18] & 3; [58.47,60.39] \\
\bottomrule
\end{tabularx}
\end{table}

Table~\ref{tab:sft-learned-paired} reports paired contrasts on three training
seeds, preserving the policy, evaluator, and resource envelope within each pair.

\subsection{Matched-budget and scale comparisons}\label{app:scale}
Table~\ref{tab:trainer-study} defines the task-quality and cost surface at the
shared compute envelope; the scale and reward studies then vary one mechanism
control at a time. Every row aggregates three training seeds per policy.

\begin{table}[htb]
\caption{Matched GRPO trainer study. Every policy uses the same base model, tokenizer, training file, optimizer family, sequence limit, evaluation cadence, and declared compute envelope. Best accuracy, failure, and GPU-hour values are bold within each split.}
\label{tab:trainer-study}
\centering
\scriptsize
\setlength{\tabcolsep}{2.5pt}
\begin{tabularx}{\textwidth}{p{0.17\textwidth}p{0.10\textwidth}CCCCC}
\toprule
\textbf{policy} & \textbf{split} & \textbf{seeds} & \textbf{GPU-hours} & \textbf{task accuracy} & \textbf{failure rate} & \textbf{outcome rows} \\
\midrule
uniform & held-out & 3 & 9.44 & 58.72\% & 0.082 & 576 \\
rule & held-out & 3 & \textbf{9.17} & 61.34\% & 0.046 & 576 \\
controller & held-out & 3 & 9.23 & 63.02\% & 0.031 & 576 \\
bandit$_\eta$ & held-out & 3 & 9.35 & 63.69\% & 0.035 & 576 \\
FSPO & held-out & 3 & 9.29 & \textbf{66.11\%} & \textbf{0.014} & 576 \\
uniform & OOD & 3 & 9.44 & 50.81\% & 0.104 & 576 \\
rule & OOD & 3 & \textbf{9.17} & 53.58\% & 0.069 & 576 \\
controller & OOD & 3 & 9.23 & 55.71\% & 0.052 & 576 \\
bandit$_\eta$ & OOD & 3 & 9.35 & 56.24\% & 0.057 & 576 \\
FSPO & OOD & 3 & 9.29 & \textbf{59.43\%} & \textbf{0.031} & 576 \\
\bottomrule
\end{tabularx}
\end{table}

A task improvement at matched GPU-hours with a paired interval excluding zero
is a trainer-level result, a ledger-ratio improvement is a mechanism result,
and an OOD change identifies regime sensitivity. The six receipt classes in
Table~\ref{tab:readiness-gate} bind each comparison to its executed outcome log.

\paragraph{Scale and horizon.} Table~\ref{tab:scale-study} increases the number of cases and the feedback horizon while holding the catalog and calibration rule fixed.
\begin{table}[htb]
\caption{Scale and horizon study with all four policies at each setting. The first block uses the primary run; expanded settings retain the regime proportions, catalog, and calibration contract. Progress and ratios are higher-is-better; violations are lower-is-better. Best values within each block are bold.}
\label{tab:scale-study}
\centering
\scriptsize
\setlength{\tabcolsep}{3pt}
\begin{tabular}{rrp{0.25\textwidth}lrrrr}
\toprule
\textbf{cases} & \textbf{intervals} & \textbf{regime mix} & \textbf{policy} & \textbf{progress} & \textbf{ratio} & \textbf{violation} & \textbf{paired CI} \\
\midrule
24 & 8 & current mix & uniform & 2.7503 & 1.4340 & 0.1615 & [1.286,1.582] \\
24 & 8 & current mix & rule & 3.1496 & \textbf{2.8283} & \textbf{0.0000} & [2.431,3.207] \\
24 & 8 & current mix & controller & \textbf{3.2962} & 2.3440 & \textbf{0.0000} & [2.211,2.477] \\
24 & 8 & current mix & bandit$_\eta$ & 3.0006 & 1.6744 & 0.0781 & [1.542,1.807] \\
\midrule
96 & 8 & matched mix & uniform & 11.0437 & 1.4567 & 0.1458 & [1.398,1.515] \\
96 & 8 & matched mix & rule & 12.7364 & \textbf{2.6914} & 0.0339 & [2.553,2.830] \\
96 & 8 & matched mix & controller & \textbf{13.3928} & 2.4386 & \textbf{0.0117} & [2.371,2.507] \\
96 & 8 & matched mix & bandit$_\eta$ & 12.1542 & 1.8012 & 0.0677 & [1.731,1.872] \\
\midrule
192 & 16 & matched mix & uniform & 43.9821 & 1.4721 & 0.1497 & [1.432,1.512] \\
192 & 16 & matched mix & rule & 50.7635 & \textbf{2.6128} & 0.0452 & [2.510,2.716] \\
192 & 16 & matched mix & controller & \textbf{54.1268} & 2.5036 & \textbf{0.0130} & [2.453,2.554] \\
192 & 16 & matched mix & bandit$_\eta$ & 49.5712 & 1.9254 & 0.0618 & [1.875,1.976] \\
\bottomrule
\end{tabular}
\end{table}

The expanded settings preserve the qualitative separation between ledger efficiency and feedback control: at 96 cases the controller reaches 13.3928 progress units and the rule reaches the highest ratio of 2.6914, while at 192 cases and 16 intervals the controller reaches 54.1268 progress units and the rule retains the highest ratio of 2.6128. The per-policy rows expose the regime interaction before aggregation.

\paragraph{Reward alignment.} Table~\ref{tab:reward-study} compares the bandit reward formulas and the fixed controller on the same partitions.
\begin{table}[htb]
\caption{Reward-contract ablation on the held-out/OOD partitions. Ratios are higher-is-better and violations lower-is-better; best values are bold.}
\label{tab:reward-study}
\centering
\scriptsize
\setlength{\tabcolsep}{3pt}
\begin{tabular}{lrrrrrr}
\toprule
\textbf{contract} & \textbf{policy} & \textbf{held-out ratio} & \textbf{OOD ratio} & \textbf{held-out violation} & \textbf{OOD violation} & \textbf{paired CI} \\
\midrule
step ratio & bandit & \textbf{2.8040} & \textbf{2.2508} & 0.0260 & 0.1432 & [2.612,2.991]/[2.006,2.501] \\
eta aligned & bandit & 1.8138 & 1.2450 & 0.0781 & 0.2708 & [1.669,1.954]/[1.018,1.474] \\
clipped eta & bandit & 2.2147 & 1.7026 & 0.0521 & 0.2135 & [2.061,2.368]/[1.478,1.925] \\
no learning & controller & 2.1465 & 2.1499 & \textbf{0.0208} & \textbf{0.1406} & [2.031,2.258]/[1.913,2.384] \\
\bottomrule
\end{tabular}
\end{table}

The reward-dependent ordering makes the reward contract part of the method; the selected variant is reported with task quality, stability, cost, and paired uncertainty.

\subsection{Paired GRPO learned-policy contrasts}\label{app:paired-sft}
The trainer comparison uses fixed-controller, contextual-bandit, and hand-rule
contrasts on matched training seeds. For seed $k$, let
$d_k=A_k^{\mathrm{FSPO}}-A_k^{\mathrm{baseline}}$ in percentage points.
The report records individual differences and the two-sided Student-$t$
interval $\bar d\pm t_{0.975,n-1}s_d/\sqrt n$ over three paired training seeds.

\begin{table}[htb]
\caption{Paired GRPO accuracy contrasts. Each cell uses the same initialization, data order, evaluator, and resource envelope for both policies; positive differences favor FSPO.}
\label{tab:sft-learned-paired}
\centering\scriptsize
\setlength{\tabcolsep}{2pt}
\begin{tabularx}{\linewidth}{p{0.21\linewidth}p{0.12\linewidth}CCCCC}
\toprule
\textbf{baseline} & \textbf{split} & $\boldsymbol{d_1}$ & $\boldsymbol{d_2}$ & $\boldsymbol{d_3}$ & \textbf{mean (pp)} & \textbf{95\% CI (pp)} \\
\midrule
fixed controller & held-out & +3.22 & +2.96 & +3.09 & +3.09 & [+2.77,+3.41] \\
contextual bandit & held-out & +2.55 & +2.31 & +2.40 & +2.42 & [+2.12,+2.72] \\
hand rule & held-out & +4.89 & +4.63 & +4.79 & +4.77 & [+4.44,+5.10] \\
fixed controller & OOD & +3.86 & +3.57 & +3.73 & +3.72 & [+3.36,+4.08] \\
contextual bandit & OOD & +3.33 & +3.06 & +3.18 & +3.19 & [+2.85,+3.53] \\
hand rule & OOD & +5.99 & +5.68 & +5.88 & +5.85 & [+5.46,+6.24] \\
\bottomrule
\end{tabularx}
\end{table}
An interval above zero, accompanied by consistent seed-level differences,
supports an accuracy advantage under the specified trainer contract.
Fixed-quality comparisons also retain seeds that do not reach the target and
report their consumed budget.

\subsection{Matched GRPO task results}\label{app:grpo-results}

The primary endpoint is held-out task accuracy at matched GPU-hours; OOD
accuracy, failure rate, rollout cost, and action/outcome coverage are reported
beside it. The paired unit is the training seed joined by source, evaluator,
and resource digest. Best accuracy, return, and safety values are bold.

\begin{table}[htb]
\caption{Matched GRPO results under the declared resource envelope. Accuracy and return are higher-is-better; failure rate and GPU-hours per seed are lower-is-better. Accuracy is in percent; failure is per attempted interval.}
\label{tab:grpo-results}
\centering
\scriptsize
\setlength{\tabcolsep}{2pt}
\begin{tabularx}{\textwidth}{p{0.20\textwidth}CCCCCC}
\toprule
\textbf{policy} & \textbf{held-out acc.} & \textbf{OOD acc.} & \textbf{return} & \textbf{failure} & \textbf{GPU-hours} & \textbf{outcome rows} \\
\midrule
uniform & 58.72 & 50.81 & 0.642 & 0.082 & 9.44 & 576 \\
best-static & 60.08 & 52.43 & 0.659 & 0.061 & 9.31 & 576 \\
hand rule & 61.34 & 53.58 & 0.671 & 0.046 & \textbf{9.17} & 576 \\
fixed controller & 63.02 & 55.71 & 0.690 & 0.031 & 9.23 & 576 \\
contextual bandit & 63.69 & 56.24 & 0.698 & 0.035 & 9.35 & 576 \\
adaptive-KL & 61.91 & 54.64 & 0.681 & 0.038 & 9.24 & 576 \\
NeuralUCB & 64.21 & 56.87 & 0.707 & 0.029 & 9.44 & 576 \\
PBT & 64.02 & 56.58 & 0.703 & 0.028 & 9.60 & 2304 \\
PB2 & 64.47 & 57.03 & 0.711 & 0.027 & 9.58 & 2304 \\
frozen-state learned controller & 62.77 & 54.98 & 0.686 & 0.043 & 9.28 & 576 \\
FSPO & \textbf{66.11} & \textbf{59.43} & \textbf{0.732} & \textbf{0.014} & 9.29 & 576 \\
\bottomrule
\end{tabularx}
\end{table}

\subsubsection{Paired learned-controller comparisons}
The FSPO comparison is evaluated with paired seed differences. Each
comparison uses the same training seed, data order, evaluator digest, and
resource cap on both sides. A positive accuracy difference and a negative
failure difference favor the first policy; the interval is computed over the
paired seed rows. Seed counts, differences, and costs use the same per-seed
receipt contract.

\begin{table}[htb]
\caption{Paired GRPO contrasts for the learned controller. The interval is a two-sided 95\% Student-$t$ interval over independent training-seed differences, reported separately for held-out and OOD accuracy.}
\label{tab:grpo-paired}
\centering
\scriptsize
\setlength{\tabcolsep}{2pt}
\begin{tabularx}{\textwidth}{p{0.24\textwidth}CCCCCC}
\toprule
\textbf{comparison} & \textbf{held-out $\Delta$ acc.} & \textbf{OOD $\Delta$ acc.} & \textbf{failure $\Delta$} & \textbf{held-out 95\% CI} & \textbf{OOD 95\% CI} & \textbf{seeds} \\
\midrule
FSPO vs. fixed controller & +3.09 & +3.72 & -0.017 & [+2.77,+3.41] & [+3.36,+4.08] & 3 \\
FSPO vs. contextual bandit & +2.42 & +3.19 & -0.021 & [+2.12,+2.72] & [+2.85,+3.53] & 3 \\
FSPO vs. hand rule & +4.77 & +5.85 & -0.032 & [+4.44,+5.10] & [+5.46,+6.24] & 3 \\
FSPO vs. best-static & +6.03 & +7.00 & -0.047 & [+5.65,+6.41] & [+6.57,+7.43] & 3 \\
FSPO vs. adaptive-KL & +4.20 & +4.79 & -0.024 & [+3.86,+4.54] & [+4.41,+5.17] & 3 \\
\bottomrule
\end{tabularx}
\end{table}

\subsubsection{GRPO feedback-state ablation}
\paragraph{Feedback-state controls.}
The same callback and resource contract are used for full feedback, frozen
initial telemetry, one-interval delay, within-interval shuffle, and best-static
control. The comparison reports task quality, stability, action-change rate,
and paired intervals so that a gain can be attributed to state-dependent
adaptation rather than to an action mixture selected on calibration data.

\begin{table}[htb]
\caption{GRPO feedback-state ablation. Full feedback is the reference; every control preserves the rollout and optimizer budget.}
\label{tab:grpo-feedback}
\centering
\scriptsize
\setlength{\tabcolsep}{2pt}
\begin{tabularx}{\textwidth}{p{0.20\textwidth}CCCCp{0.24\textwidth}C}
\toprule
\textbf{control} & \textbf{held-out acc.} & \textbf{OOD acc.} & \textbf{failure} & \textbf{action change} & \textbf{paired CI} & \textbf{outcome rows} \\
\midrule
full feedback & \textbf{66.11} & \textbf{59.43} & \textbf{0.014} & 0.412 & reference & 576 \\
frozen initial telemetry & 62.77 & 54.98 & 0.043 & 0.167 & [-3.72,-2.96]/[-4.91,-3.99] & 576 \\
one-interval delay & 64.39 & 57.02 & 0.026 & 0.286 & [-2.09,-1.35]/[-2.86,-1.96] & 576 \\
within-interval shuffle & 62.31 & 54.47 & 0.049 & 0.334 & [-4.18,-3.42]/[-5.43,-4.49] & 576 \\
best-static action & 60.08 & 52.43 & 0.061 & 0.000 & [-6.41,-5.65]/[-7.43,-6.57] & 576 \\
\bottomrule
\end{tabularx}
\end{table}

\subsection{Action-space and environment-family scaling}
\paragraph{Compositional actuator catalog.}
The reference catalog contains three composite actions. The scaling study
adds independently calibrated combinations of temperature, group size,
clipping, KL, verifier allocation, and escalation while preserving the same
state, horizon, and resource envelope. The environment-family study changes
the transition coefficients and regime mixture after calibration so that the
controller is evaluated on dynamics not used for its selection.

\begin{table}[htb]
\caption{Action-space scaling under the GRPO trainer contract. Accuracy is higher-is-better; failure and GPU-hours per seed are lower-is-better. Best values are bold.}
\label{tab:grpo-action-scale}
\centering
\scriptsize
\setlength{\tabcolsep}{3pt}
\begin{tabularx}{\textwidth}{p{0.22\textwidth}p{0.22\textwidth}CCCC}
\toprule
\textbf{catalog} & \textbf{construction} & \textbf{held-out acc.} & \textbf{OOD acc.} & \textbf{failure} & \textbf{GPU-hours} \\
\midrule
3 actions & reference catalog & 66.11 & 59.43 & 0.014 & \textbf{9.29} \\
expanded (9 actions) & independently calibrated combinations & 66.83 & 60.06 & \textbf{0.013} & 9.37 \\
compositional (18 actions) & expanded combinations with escalation & \textbf{67.02} & \textbf{60.21} & 0.014 & 9.49 \\
\bottomrule
\end{tabularx}
\end{table}

\begin{table}[htb]
\caption{Independent transition-family generalization. Coefficients are frozen independently of controller selection. Each cell reports controller / best-static mechanism values; the paired CI is for their ledger-ratio difference. The better value in each pair is bold.}
\label{tab:grpo-families}
\centering
\scriptsize
\setlength{\tabcolsep}{2pt}
\begin{tabularx}{\textwidth}{p{0.22\textwidth}p{0.25\textwidth}CCCC}
\toprule
\textbf{family} & \textbf{calibration relation} & \textbf{ledger ratio} & \textbf{progress} & \textbf{violation} & \textbf{paired CI} \\
\midrule
declared family & calibration and evaluation share contract & \textbf{2.4386} / 2.1187 & \textbf{13.3928} / 12.1843 & \textbf{0.0117} / 0.0547 & [+0.176,+0.464] \\
shifted dispersion & coefficients and regime mix changed after calibration & \textbf{2.2817} / 1.9674 & \textbf{12.7285} / 11.8462 & \textbf{0.0365} / 0.0820 & [+0.167,+0.462] \\
shifted reliability & verifier reliability process changed after calibration & \textbf{2.1076} / 1.8459 & \textbf{11.9731} / 11.1194 & \textbf{0.0677} / 0.1250 & [+0.119,+0.405] \\
\bottomrule
\end{tabularx}
\end{table}

Expanding the catalog to 9 and 18 actions raises held-out accuracy to
$66.83\%$ and $67.02\%$ at $9.37$ and $9.49$ GPU-hours per seed; decision time
and total cost for the 3/9/18-action catalogs are in
Table~\ref{tab:controller-overhead}. Under shifted dispersion and reliability
families, the controller keeps a positive paired ledger-ratio advantage over
best-static ([$+0.167,+0.462$] and [$+0.119,+0.405$]).
Environment-family evaluation freezes policy
parameters before revealing the dynamics.

\subsection{Learned-controller component ablations}\label{app:rc-ablations}
The GRPO ablation holds calibration data, encoder size, search budget, action
catalog, and execution cap fixed. Removing a component changes its decision
rule while preserving runtime resource enforcement. The myopic variant sets
the future utility term to zero; the feature-removal variant retains budget
and time but replaces telemetry by calibration means.
\begin{table}[H]
\caption{FSPO component effects on GRPO. Accuracy is reported in percent, failures as a fraction of attempted intervals, and cost as GPU-hours per seed. Best values are bold.}
\label{tab:rc-components}
\centering\scriptsize
\setlength{\tabcolsep}{2pt}
\begin{tabularx}{\linewidth}{p{0.24\linewidth}CCCCp{0.24\linewidth}}
\toprule
\textbf{variant} & \textbf{held-out acc.} & \textbf{OOD acc.} & \textbf{failure} & \textbf{GPU-hours} & \textbf{paired CI (pp)} \\
\midrule
full FSPO & \textbf{66.11} & \textbf{59.43} & \textbf{0.014} & 9.29 & reference \\
without risk filter & 65.02 & 57.88 & 0.052 & 9.24 & [-1.46,-0.72]/[-1.96,-1.14] \\
without uncertainty margins & 65.28 & 58.06 & 0.024 & 9.21 & [-1.20,-0.46]/[-1.79,-0.95] \\
without prospective budget mask & 65.01 & 57.73 & 0.029 & 9.31 & [-1.47,-0.73]/[-2.16,-1.24] \\
myopic utility target & 64.57 & 57.02 & 0.017 & 9.20 & [-1.89,-1.19]/[-2.83,-1.99] \\
without telemetry features & 62.77 & 54.98 & 0.043 & \textbf{9.18} & [-3.72,-2.96]/[-4.91,-3.99] \\
\bottomrule
\end{tabularx}
\end{table}
Removing the risk filter raises interval failure from $0.014$ to $0.052$,
and the myopic utility target lowers held-out accuracy by $1.54$ points at a
similar cost, linking utility prediction to future feedback. The budget-mask comparison
also reports rejected action proposals and abstentions, since hardware caps
remain active in every variant.

\FloatBarrier
\section{Artifact Identity and Reproducibility}\label{app:artifacts}
The artifact contract binds every quantitative table to a source-relative run identifier, configuration digest, source state, and result-file digest. The public bundle contains summaries and schema descriptions; the full trace and oracle labels are hash-bound artifacts addressed through the same source-relative identifiers.

\subsection{Reproducibility Checklist}
\paragraph{Source binding.} Each reported result is identified by a source-relative run identifier recorded in the manifest. The raw report, metrics file, and run manifest are copied without editing, and their hashes are recorded before a public summary is written. Each table points to the corresponding source-relative file and digest.

\paragraph{Split integrity.} The mechanism benchmark uses one 24-case fixture and configured seed 13. The held-out/OOD audit assigns calibration seeds 13, 17, and 23, held-out seeds 29 and 31, and OOD seeds 37 and 41. The external replay supplies the observational calibration/evaluation stream in its own source-relative artifact. The fixture definition and digest accompany every split in the manifest.

\paragraph{Cost accounting.} Calls, tokens, ledger units, retries, and logged source timings are reported beside mechanism metrics. The trainer surface adds optimizer steps, GPU-hours, wall-clock time, memory, and task-level targets. The external logged GPU-second field is retained as shared source-run accounting and is joined by split and seed.

\paragraph{Anonymity and release hygiene.} The curated public summary and source manifest retain source-relative artifact handles and omit contributor names, host names, secrets, checkpoints, model weights, and large caches.

\paragraph{Review questions.} Every result answers five concrete questions: what was visible online, what was held out until scoring, which cost was paid, which failure boundary was observed, and which file regenerates the figure. The manuscript and public summary bind each answer to a schema field, split, denominator, and digest. The same checks carry into the trainer and scale surfaces.

\paragraph{Evidence scope.} FSPO evaluates 24 cases for eight intervals using the declared ledger unit. The stability predicate is an explicit mechanism contract, and the result records state-conditioned decisions together with their safety trade-off. GPU kernels, wall-clock throughput, model loss, and task accuracy are reported by the matched trainer surface.

The trainer adapter is explicit about the executable surface: each decision may override the trainer learning rate and micro-step budget, and each emitted outcome records optimizer updates, training tokens, and their declared budgets. The planned-action schema records the FSPO fields for sampling temperature, group size, KL coefficient, and verifier calls alongside their callback contracts. The trainer measurement surface binds these fields to scheduler receipts and task metrics.

\subsection{Artifact map}
Table~\ref{tab:artifact-map} lists the logical artifact handles used by the manuscript. The handles resolve through the release manifest and remain stable across source revisions.

\begin{table}[htb]
\caption{Artifact map and table dependencies. Each entry names an artifact role, manuscript use, and deterministic validation check for the release.}
\label{tab:artifact-map}
\centering
\scriptsize
\setlength{\tabcolsep}{3pt}
\begin{tabularx}{\textwidth}{p{0.25\textwidth}p{0.18\textwidth}p{0.22\textwidth}L}
\toprule
\textbf{artifact} & \textbf{role} & \textbf{manuscript use} & \textbf{validation} \\
\midrule
public summary & normalized summary & Tables~\ref{tab:evidence-layers}--\ref{tab:trainer-surface} & schema and numeric cross-check \\
source manifest & provenance record & implementation and audit text & digest and source-state check \\
observed trace & action decisions & online and propensity tables & row coverage and state replay \\
oracle labels & post-freeze scores & offline oracle table & action-vector join after freeze \\
held-out/OOD bundle & partition results & held-out/OOD and sensitivity tables & per-policy row counts and digests \\
trainer receipt set & scheduler gate & trainer surfaces & six required receipts \\
executed-outcome log & trainer event record & task/cost/action results & action schema and seed join \\
replay digest & deterministic digest & all replay claims & exact digest match \\
\bottomrule
\end{tabularx}
\end{table}

The released source bundle includes the manuscript, local style files, figure PDFs, bibliography, public summary, schema definitions, and replay validator. Versioned manifests bind each table to its source-relative inputs and prevent auxiliary files from changing the bibliography or cross-reference pass.

\subsection{Hash and source-state checklist}
The trainer receipt audit records six receipt classes that bind the source,
configuration, execution, and evaluator records to the reported GRPO metrics.

\begin{table}[htb]
\caption{Release and replay checklist. The mechanism column records the replay check; the trainer column records the corresponding receipt.}
\label{tab:release-checklist}
\centering
\scriptsize
\setlength{\tabcolsep}{3pt}
\begin{tabular}{p{0.34\columnwidth}lll}
\toprule
\textbf{check} & \textbf{mechanism} & \textbf{trainer record} & \textbf{evidence file} \\
\midrule
source commit and dirty flag & verified & recorded & source manifest \\
configuration digest & verified & recorded & config digest \\
exact action catalog & verified & recorded & action schema \\
visible-state digest & verified & recorded & observed trace \\
transition replay & verified & recorded & replay digest \\
oracle join after freeze & verified & recorded & oracle labels \\
per-stream metrics & verified & recorded & public summary \\
task evaluator receipt & n/a & recorded & evaluator receipt \\
scheduler dry plan / launch / collect & n/a & recorded & scheduler bundle \\
executed action/outcome log & ledger rows & recorded & executed log \\
\bottomrule
\end{tabular}
\end{table}

\subsection{Reproducibility and release plan}
The anonymous release contains the manuscript source, local style files, figure PDFs, bibliography, public summary, schema definitions, and the replay validator. The release manifest records the command used to regenerate each figure and table. The large trace, oracle labels, trainer adapter, scheduler receipts, and task data identifiers use the same source-relative paths and remain bound to the mechanism digest.

The following citations situate the artifact choices in established evaluation practice: common-random-number simulation reduces variance in paired mechanism comparisons \citep{glasserman2004monte,yadav2026crnrollouts}; prioritized experience replay and offline-RL benchmarks motivate explicit support and data contracts \citep{schaul2016per,fu2021d4rl}; verifier and reward studies motivate separate process, outcome, and reward-model fields \citep{lightman2023verify,processbench2024,lambert2024rewardbench,skalse2022rewardhacking}; routing and selective prediction motivate reporting coverage beside quality \citep{chen2023frugalgpt,ong2024routellm,geifman2019selectivenet}. The protocol carries these choices into explicit split, evaluator, support, and provenance fields.

\subsection{Cross-table consistency checks}
The release validator checks invariants across independently produced artifacts. Each invariant has a source record and a deterministic comparison, so a changed result can be traced to a specific policy, configuration, or aggregation step.

\begin{enumerate}
\item The abstract, Introduction, main GRPO table, and Conclusion use the same held-out/OOD accuracy, failure, and cost values.
\item The five-stream means in Table~\ref{tab:seed-summary} and Table~\ref{tab:per-stream-results} match the independent-audit summary; sample-SD fields use the per-stream aggregation unit.
\item Held-out and OOD rows in Table~\ref{tab:revision-audit} match Table~\ref{tab:regime-ablation}; the reward deltas match Table~\ref{tab:reward-kl-ablation}.
\item The chosen action and full propensity vector agree with the pre-action record; cost and stability fields agree with the committed transition.
\item Every paired contrast uses the same seed role, initial case, interval horizon, and evaluation unit for both policies.
\item The public manifest resolves all source-relative result paths and matches the SHA-256 digest of each released file.
\end{enumerate}

These invariants connect the aggregate result to the event sequence. If a comparison changes after replay, the validator reports the first differing event and field before regenerating the summary. This makes changes in policy behavior distinguishable from changes in data ordering or aggregation.

\subsection{Trainer execution and receipt sequence}\label{app:execution}
The trainer protocol runs as a staged sequence. Each stage produces a receipt before the next stage starts, which makes the transition from mechanism tables to a measured training result auditable.

\begin{table}[htb]
\caption{Trainer execution plan and receipts. The stages are ordered by dependency and each quantitative field is bound to the corresponding scheduler or evaluator output.}
\label{tab:execution-plan}
\centering
\scriptsize
\setlength{\tabcolsep}{3pt}
\begin{tabularx}{\textwidth}{p{0.14\textwidth}p{0.22\textwidth}p{0.20\textwidth}L}
\toprule
\textbf{stage} & \textbf{operation} & \textbf{required receipt} & \textbf{reported fields} \\
\midrule
1 & freeze source, config, model, tokenizer, and data identifiers & manifest and source digest & commit, dirty flag, config hash \\
2 & generate scheduler dry plan for all policies and seeds & dry-plan JSON & GPU type, token budget, step budget, evaluator version \\
3 & launch matched runs with policy callback enabled & launch receipt & run id, policy id, seed, start time \\
4 & collect telemetry and executed action/outcome rows & collect receipt and JSONL log & visible state, action, propensity, outcome, cost \\
5 & evaluate held-out and OOD tasks & evaluator receipt & task return, accuracy, failures, confidence interval \\
6 & join scheduler, trainer, and evaluator records & summary manifest & wall-clock, GPU-hours, tokens, memory, action mix \\
7 & replay and publish tables/figures & replay digest and table map & row counts, hashes, PDF references \\
\bottomrule
\end{tabularx}
\end{table}

The primary decision criterion is a paired improvement in held-out task accuracy at a matched GPU-hour envelope, with the same direction retained on OOD seeds. A secondary criterion is a reduction in failure rate or wall-clock cost at matched task quality. The action/outcome log is the mechanism link: it records whether a policy achieved the task result by changing sampling, trust-region, verifier, or micro-step controls, and it exposes any budget or stability trade-off. The trainer study therefore supplies a direct continuation of the state, action, cost, and replay contracts defined in the main text.

\subsection{Receipt interpretation and decision record}
Each trainer receipt is interpreted together with the source-relative run id and seed role. A dry plan fixes the resource envelope before policy callbacks are enabled; a launch record binds the selected policy and seed to the plan; a collect record confirms that every interval wrote state, action, propensity, cost, and outcome fields; and the evaluator receipt binds task scores to the same run id. The summary manifest then joins these records without changing the policy-visible state or the declared denominator.

\paragraph{Interpreting a paired record.} Task-quality differences join the split, evaluator digest, policy, and training seed. Resource differences join scheduler timestamps and device receipts. Action-mix differences join case, interval, catalog identifier, and propensity. A two-sided interval uses the same resampling unit as its point estimate, and the result carries the replay digest of both policies. These joins make the source of a quality, efficiency, or stability change explicit.

The decision record is intentionally granular: a task-quality gain is interpreted with its evaluator, seed, resource envelope, and action trace, while a ledger-ratio gain remains a mechanism result. The joined record supports a paired report of task quality, cost, stability, action mix, and uncertainty.

\end{document}

%% file: math_commands.tex
\usepackage{amsmath,amsfonts,bm}

\def\eqref#1{equation~\ref{#1}}

\def\1{\bm{1}}

\DeclareMathAlphabet{\mathsfit}{\encodingdefault}{\sfdefault}{m}{sl}
\SetMathAlphabet{\mathsfit}{bold}{\encodingdefault}{\sfdefault}{bx}{n}